%% file: main.tex
\pdfoutput=1
\documentclass[11pt]{article}

\usepackage[final]{ACL2023}
\usepackage{times}
\usepackage{latexsym}
\usepackage{pgffor}
\usepackage[T1]{fontenc}
\usepackage[utf8]{inputenc}
\usepackage{tikz}
\usepackage{tikzit}
\input{sample.tikzstyles}
\tikzset{
    state/.style={
        circle,
        draw=black,
        minimum size=1cm,
    },
    arrow/.style={
        ->,
        thick,
        >=stealth,
    },
    label/.style={
        sloped,
        above
    }
}
\usepackage{microtype}

\usepackage{inconsolata}
\usepackage{xcolor}
\definecolor{cadmiumgreen}{rgb}{0.0, 0.8, 0.08}
\usepackage{xspace}

\newcommand{\parentDatasetName}{{\sc InfoTabS}\xspace}
\newcommand{\contra}{{\textcolor{red}{\textbf{C}}}\xspace}
\newcommand{\neutral}{{\textcolor{gray}{\textbf{N}}}\xspace}
\newcommand{\entailment}{{\textcolor{cadmiumgreen}{\textbf{E}}}\xspace}
\usepackage{booktabs}
\usepackage{amsmath}
\usepackage{rotating}
\usepackage{subcaption}
\usepackage{tabularx}
\usepackage{multirow,longtable}
\usepackage{diagbox} % Add the diagbox package
\usepackage{listings}

\newcommand{\bert}{{\sc BERT}\xspace}
\newcommand{\roberta}{{\sc RoBERTa}\xspace}
\newcommand{\robertaL}{{\sc RoBERTa-Large}\xspace}
\newcommand{\model}{{\sc RoBERTa$_{\text{\sc InTa}}$}\xspace}

\usepackage{paralist}
\usepackage{tcolorbox}
\usepackage{mdframed}

\newcommand{\exampleParagraph}[2]{\framebox{\parbox{0.98\linewidth}{\paragraph{#1} 
{\footnotesize #2}}}}

\definecolor{darkteal}{rgb}{0, 0.5, 0.5}
\title{{\small \hfill EMNLP'24}\\
\vspace*{.25in} Evaluating Concurrent Robustness of Language Models \\ Across Diverse Challenge Sets}

\author{
Vatsal Gupta\textsuperscript{\rm 1†},
Pranshu Pandya\textsuperscript{\rm 1†},
Tushar Kataria\textsuperscript{\rm 2},
Vivek Gupta\textsuperscript{\rm 3}\thanks{~~Corresponding Author (work done while at UPenn), †Equal Contribution}~,
Dan Roth\textsuperscript{\rm 4}\\ 
\textsuperscript{\rm 1}IIT Guwahati,
\textsuperscript{\rm 2}University of Utah,
\textsuperscript{\rm 3}Arizona State University,
\textsuperscript{\rm 4}University of Pennsylvania,\\
\small {\{g.vatsal,p.pandya\}@iitg.ac.in, tkataria@cs.utah.edu, vgupt140@asu.edu, danroth@seas.upenn.edu } \\
}

\begin{document}
\maketitle
\begin{abstract}
Language models, characterized by their black-box nature, often hallucinate and display sensitivity to input perturbations, causing concerns about trust. To enhance trust, it is imperative to gain a comprehensive understanding of the model's failure modes and develop effective strategies to improve their performance. In this study, we introduce a methodology designed to examine how input perturbations affect language models across various scales, including pre-trained models and large language models (LLMs). Utilizing fine-tuning, we enhance the model's robustness to input perturbations. Additionally, we investigate whether exposure to one perturbation enhances or diminishes the model's performance with respect to other perturbations. To address robustness against multiple perturbations, we present three distinct fine-tuning strategies. Furthermore, we broaden the scope of our methodology to encompass large language models (LLMs) by leveraging a chain of thought (CoT) prompting approach augmented with exemplars. We employ the Tabular-NLI task to showcase how our proposed strategies adeptly train a robust model, enabling it to address diverse perturbations while maintaining accuracy on the original dataset. 
% \textit{Code and Data to be released upon acceptance.}
\end{abstract}

\section{Introduction}

\input{introduction}

\section{Proposed Methodology} 

\input{Methodology}

\section{Case Study on Tabular Inference}
\input{Experiments}

\section{Results and Analysis}
\input{Results}

% \vspace{-0.5em}
\section{Related Works}
% \vspace{-1em}
\input{related-works}

\section{Conclusion and Future Works}
  We demonstrate that input perturbation poses difficulties for LMs at all scales. While fine-tuned models on a single challenge set can produce robust models, their generalizability to unfamiliar perturbations remains questionable. This motivates the problem of multi-set inoculation, aiming to train a singular model resilient to a myriad of distinct perturbations. We introduce a comprehensive framework to systematically evaluate LM robustness against multiple input perturbations. Additionally, we propose three strategies to fine-tune the model on multiple challenge/pertubations sets. Our results underscore the superiority of mixed fine-tuning in training robust models. Furthermore, we expand our framework to LLMs, leveraging a \textit{COT} prompting enriched with exemplar demonstrations.
 %\vspace{-0.75em}

\textbf{Future Directions:} 
We consider the following future directions:
\begin{inparaenum}[(a.)] \item \textbf{Complex Sample Selection:} Future plans include adopting advanced sample selection strategies to boost model robustness during fine-tuning, inspired by \citet{roh2021sample, swayamdipta2020dataset}.\item \textbf{Composite Perturbation:} We aim to explore the successive application of multiple perturbations on a single sample, represented as $\pi_i(\pi_j(x))$, to understand their combined impact on model performance. 
\end{inparaenum}

\section*{Limitations}
While our framework exhibits promising results for language models at different scales, there are several limitations to consider. We study five different perturbations in our framework. The effectiveness of our method, however, is contingent on the availability of data and definitions of these perturbations, which may not be available for unique unencountered perturbations. In addition, the process of sequential fine-tuning presents a challenge in terms of catastrophic forgetting. This necessitates maintaining a repository of both current and historical data and perturbations, which in turn leads to an increase in computational storage. Although our system performs well for tasks in English, processing and adapting to multilingual input data and accompanying models is an area that has to be researched further. We also recognize the opportunity for investigating parameter-efficient fine-tuning and other domain adaptation strategies to potentially enhance the robustness of the model. Finally, it is pertinent to note that the current evaluation of our framework has been limited to specific natural language processing tasks. Its performance in other tasks, such as question-answering and sentiment classification, has not yet been explored. These limitations underscore the need for further research to address these challenges.

\section*{Ethics Statement}
We, the authors of this work, affirm that our work complies with the highest ethical standards in research and publication. In conducting this research, we have considered and addressed various ethical considerations to ensure the responsible and fair use of computational linguistics methodologies.  We provide detailed information to facilitate the reproducibility of our results. This includes sharing code, datasets (in our case, we deal with publicly available datasets and comply with the ethical standards mentioned by the authors of the respective works.), and other relevant resources to enable the research community to validate and build upon our work. The claims in the paper match the experimentation results. However, a certain degree of stochasticity is expected with \textit{black-box} large language models, which we attempt to minimize by keeping a fixed temperature. We describe in the fullest detail the annotations, dataset splits, models used, and prompting methods tried, ensuring the reproducibility of our work. For grammar correction, we use AI-based writing assistants, and for coding, we utilized Copilot. It's important to note that the genesis of our ideas and the conduct of our research were entirely independent of AI assistance.

\section*{Acknowledgements}
Research was sponsored by the Army Research Office and was accomplished under Grant Number
W911NF-20-1-0080. The views and conclusions contained in this document are those of the authors and should not be interpreted as representing the official policies, either expressed or implied, of the Army Research Office or the U.S. Government. The U.S. Government is authorized to reproduce and distribute reprints for Government purposes notwithstanding any copyright notation herein. This work was partially funded by ONR Contract N00014-23-1-2364. We extend our gratitude to the annotators who verified our flowcharts and corresponding question answer pairs. Lastly, we extend our appreciation to the reviewing team for their insightful comments.

\bibliography{anthology,custom}
\bibliographystyle{acl_natbib}
\appendix
\section{Appendix}
\input{Appendix}

\end{document}

%% file: sample.tikzstyles
\tikzstyle{new style 0}=[fill=white, draw={rgb,255: red,173; green,175; blue,191}, shape=circle, minimum width=.5cm, ultra thick]
\tikzstyle{new style 1}=[fill=white, draw=black, shape=rectangle]
\tikzstyle{new style 2}=[fill=white, draw=red, shape=rectangle]

\tikzstyle{new edge style 0}=[->, fill=none, draw=red, ultra thick]
\tikzstyle{new edge style 1}=[<-, fill=none, ultra thick]
\tikzstyle{new edge style 2}=[draw=black, fill=none, dashed, <-, ultra thick]
\tikzstyle{new edge style 3}=[<-, ultra thick]
\tikzstyle{new edge style 4}=[->, ultra thick]
\tikzstyle{new edge style 5}=[draw=black, fill=none, dashed, ->, ultra thick]

%% file: introduction.tex
Language models (LMs), which have become increasingly integrated into various aspects of daily lives, hold immense potential to revolutionize how we interact with technology. Their ubiquity underscores the importance of thoroughly examining their robustness and generalizability, which will be instrumental in fostering trust among users. One notable challenge is their sensitivity to even slight changes in input. For instance, while a human can easily interpret and understand a statement regardless of minor alterations, LMs struggle \citep{wang2023adversarial, nie-etal-2020-adversarial}. This inconsistency becomes notably apparent when minor perturbations to the input, which do not inherently modify the underlying meaning, result in a marked decline in the performance of the model \citep{shankarampeta2022enhancing,glockner-etal-2018-breaking}. Examples of such perturbations for the task of tabular inference \citet{gupta-etal-2020-infotabs}, is illustrated in Figure ~\ref{fig:Sample}.

\begin{figure} 
\centering 
\begin{subtable}[b]{0.45\textwidth} 
\centering
\renewcommand{\arraystretch}{1}
\footnotesize{
\begin{tabular}{l l}\\
\hline
\multicolumn{2}{c}{\textbf{Case Closed}} \\
\hline
\textbf{Written} & Takahiro Arai \\
\textbf{Publish} & Shogakukan \\
\textbf{Eng. Publish} & SG Shogakukan Asia \\
\textbf{Demographic} & Shonen \\
\textbf{Magazine} & Weekly Shonen Sunday \\
\textbf{Orig. Run} & May 9, 2018 - present \\
\textbf{Volumes} & 2 (List of volumes) \\
\hline
\end{tabular}}
%\caption{Case Closed Details} 
\label{tab:case_details}
\end{subtable}
\hfill 
\begin{subtable}[b]{0.45\textwidth}  
\centering
\footnotesize
\fbox{\begin{minipage}{23em}
\textbf{H$_1$:} Takahiro Arai wrote `Case Closed' comic series. \textcolor{cadmiumgreen}{(E)} \\
\textbf{H$_1^{'}$:} Takahiro Arai wotte `Case Closed' comci series. \textcolor{cadmiumgreen}{(E)}
\\
\textbf{H$_2$:} `Case Closed' is a long-term comic series.\textcolor{cadmiumgreen}{(E)}
\\
\textbf{H$_2^{'}$:}`Case Closed' isn't a long-term comic series.\textcolor{red}{(C)} \\
\textbf{H$_3$:}`Case Closed' became the anime Detective Conan \textcolor{gray}{(N)} \\
\textbf{H$_3^{'}$:}Detective Conan is `Case Closed' anime version. \textcolor{gray}{(N)} \\
\textbf{H$_4$:}`Case Closed' has run over 5 years.\textcolor{cadmiumgreen}{(E)} \\
\textbf{H$_4^{'}$:}`Case Closed' has run over 10 years.\textcolor{red}{(C)} \\
\textbf{H$_5$:} Shogakukan Asia published `Case Closed' (Eng). \textcolor{cadmiumgreen}{(E)}\\ 
\textbf{H$_5^{'}$:}Shogakukan UK published `Case Closed' (Eng). \textcolor{red}{(C)}
\end{minipage}}
%\caption{Hypotheses}
\label{tab:hypotheses}
\end{subtable}
\vspace{-0.5em}
\caption{\textbf{An example of tabular premise and hypotheses from \parentDatasetName~ \citep{gupta-etal-2020-infotabs}.} Original hypotheses (H$_1$,H$_2$,H$_3$,H$_4$,H$_5$) and perturbed hypothesis (H$^{'}_{1}$,H$^{'}_{2}$,H$^{'}_{3}$,H$_{4}^{'}$,H$_{5}^{'}$) representing character, negation, paraphrasing, numeric and location perturbations respectively. Labelled as \textcolor{cadmiumgreen}{E}ntailment, \textcolor{red}{C}ontradiction or \textcolor{gray}{N}eutral. The \textbf{bold} entries in the first column are the keys, and the corresponding entries in the second column are their values.}
\label{fig:Sample} 
\vspace{-1.75em}
\end{figure}

\begin{figure*}[h]
    \centering
\includegraphics[width=1\linewidth]{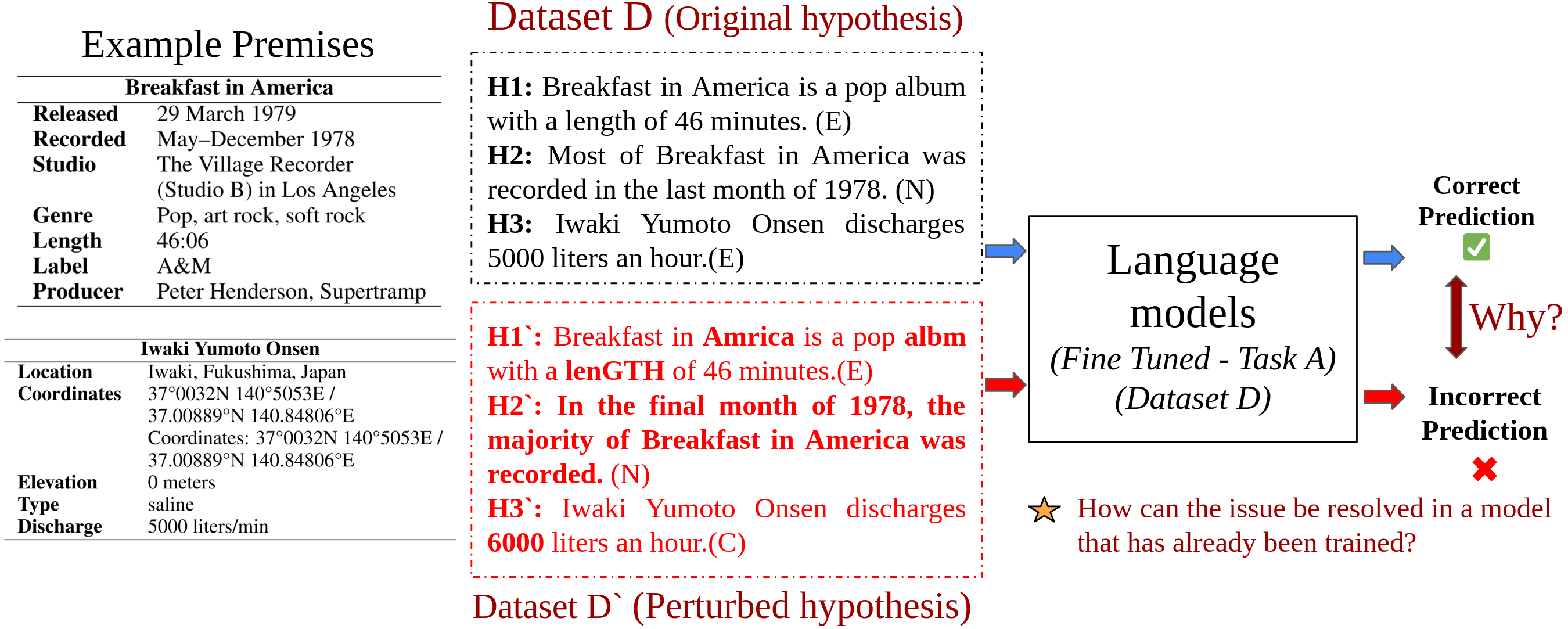}
    \caption{\textbf{Language Models Sensitivity to Input Perturbations.} Language models trained on Tabular-NLI (\textit{Task A}) with Original Hypothesis (Dataset D) are not reliable for perturbed hypotheses shown in Dataset D' for character, paraphrasing, or numeric perturbation examples.}
    \label{fig:examples}
    \vspace{-0.75em}
\end{figure*}
Addressing these sensitivities to input perturbation is crucial for the advancement and reliability of LMs in real-world applications.
Empirical evidence supports the effectiveness of fine-tuning models using perturbed input samples from challenge sets  \citep{jiang-etal-2022-rose,fursov2021differentiable}. 
For instance, \citet{wang-etal-2020-cat,liu-etal-2019-inoculation} showcased that a pre-trained language model (PLM) utilizing Masked Language Modeling (MLM) and trained for a specific NLP task becomes significantly robust to input perturbations when further fine-tuned using a small set of perturbed examples. However, the ability of these models to generalize across different types of perturbations is still a subject of investigation \citep{liu2020adversarial}. The implications of fine-tuning a model on a particular challenge/perturbation set, especially concerning its impact on handling other perturbations, warrant further exploration (refer to Figure \ref{fig:examples}). It remains unclear if a model's increased robustness to character perturbations post-fine-tuning extends to addressing challenges from other perturbations, like paraphrasing.

In this study, we address LMs robustness to input perturbations, seeking to answer the following two questions: \emph{How does fine-tuning a model on one perturbation set affect performance on other types of perturbations? Is it possible to guarantee consistent robustness across multiple distinct perturbation sets?} In particular, we extend the \emph{single-set inoculation} approach of \citet{liu-etal-2019-inoculation}, to a more generic multi-sets robustness, which we refer to as \emph{multi-set inoculation}. 
To the best of our knowledge, we are the first to introduce and extensively study the robustness of LMs to multiple perturbations.

Our proposed methodology is adept at handling both (a) transformer-based pre-trained language models (PLMs) such as \bert \citep{devlin2018bert} and \roberta\citep{liu2019roberta} 
, which are amenable to direct fine-tuning on end-user GPUs, and (b) large generative language models such as gpt-3.5-turbo (GPT-3.5) \citep{brown2020language}, GPT-4, and LLaMA, LLaMA-2 \citep{touvron2023llama}, Flan-T5 \citep{chung2022scaling,kanakarajan-sankarasubbu-2023-saama}, which are costly and have limited access to re-training (and model weights). For these generative models, we leverage the few-shot Chain of Thought \citep{wei2023chainofthought} 
as an alternative to traditional fine-tuning. This methodology circumvents the computational intricacies inherent in the fine-tuning of LLMs. It proficiently manages the tuning of a multitude of model parameters using a limited constrained set of training samples. %\tk{Kind fo repeated this 'To the best of our knowledge...' twice, we can merge the paras maybe} 
Additionally, we also study Inoculation with LLM, prior studies \citet{liu2019roberta,wang2021infobert, liu2019inoculation}  have been limited to traditional BERT style models.
 Within our framework, we investigate three distinct multi-set fine-tuning methods for PLMs and adapt them for LLMs via COT, each designed to assess and enhance model robustness across diverse perturbation sets. 
Our study makes the following contributions:

\begin{itemize}
\vspace{-0.5em}
\setlength\itemsep{-0.5em}
    \item We introduce \textit{Multi-set Inoculation}, which examines the implications of fine-tuning across multiple perturbation sets. We assess three unique multi-set fine-tuning approaches, each showing concurrent robustness to multiple perturbation sets. 
    \item We evaluate the efficacy of our framework across a spectrum of models, ranging from traditional pre-trained language models (PLMs) like RoBERTa to expansive large language models (LLMs) such as GPT-3.5 and LLaMA-2, among others, in the context of the Tabular NLI task.
\end{itemize}

Code and dataset for the experiments with Multi- set Inoculation framework on different models are available at: \href{https://msin-infotabs.github.io/}{https://msin-infotabs.github.io/}.

%% file: Methodology.tex
% \vspace{-0.5em}
\begin{figure*}[h]
    \centering
\includegraphics[width=0.79\linewidth]{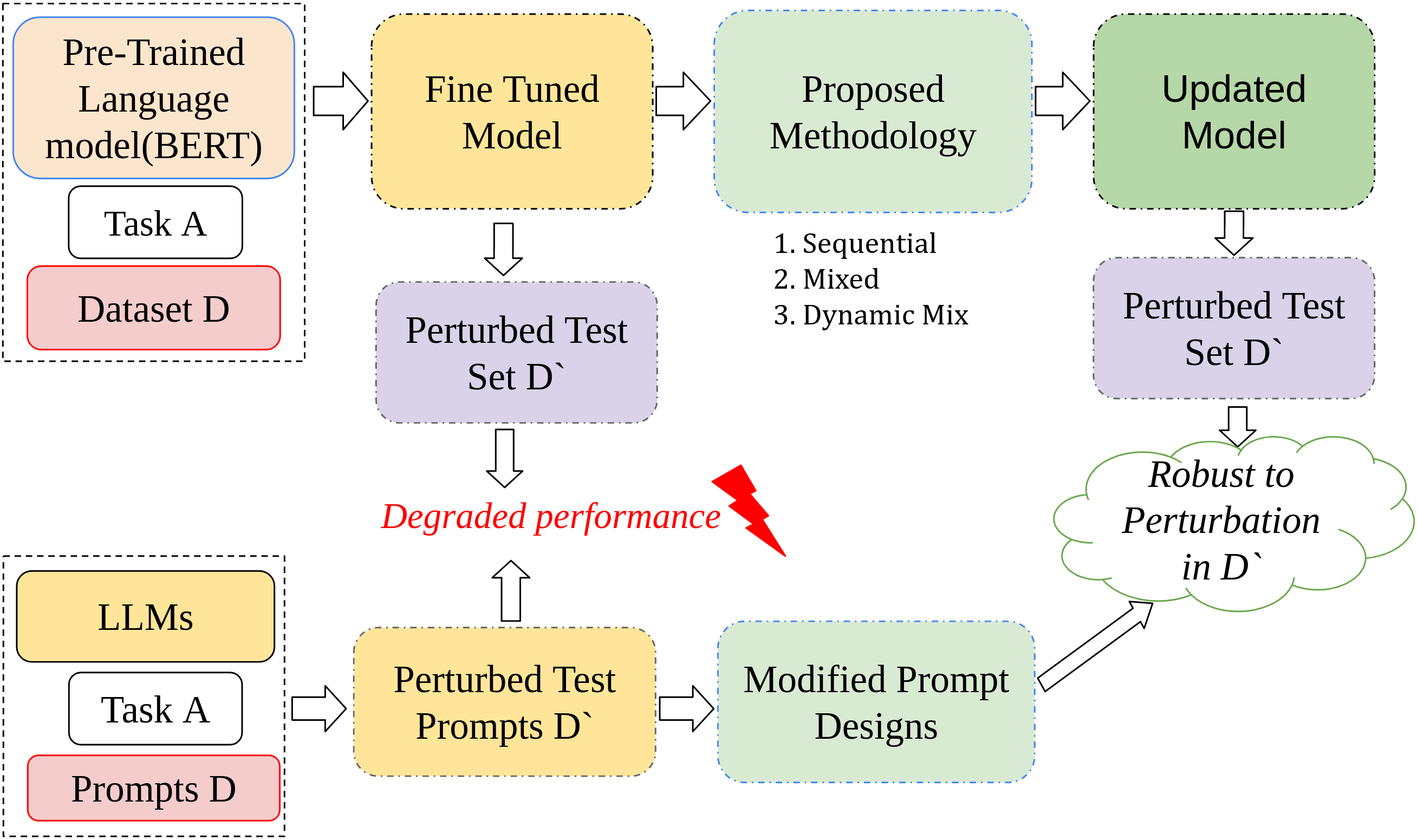}
% \vspace{-0.4em}
    \caption{\textbf{Multi-Set Inoculation Framework.} High-level flowchart describing the proposed frameworks for PLMs (via fine-tuning) and LLMs (via prompt design).}
    \label{fig:flowchart}
    \vspace{-1.25em}
\end{figure*}
In this section, we detail the methodology for \textit{Multiset Inoculation}. We evaluate the robustness of the model by subjecting it to different input perturbations. Subsequently, we introduce multiset fine-tuning techniques, which improve the model's performance on diverse perturbed datasets. Figure \ref{fig:flowchart} shows a high-level flowchart of our methodology.

%\tk{We should consider seperating out formulae into new line instead of paragraph}
%\tk{Shouldn't we define what is xi, yi here }
% \textbf{Terminology.} Given a pre-trained language model (PLM) denoted as {\sc{M}}, fine-tuned on the original (unperturbed) training set {\sc O} $= \{(x_i, y_i)\}_{i=1}^{N}$ for a natural language processing (NLP) task {\sc T}. Let \{{\sc $\pi_j$}\}$_{j=1}^{m}$ represent input perturbations, where $m$ is the number of distinct perturbations available. For each perturbation $j$, let O${\sc _{S_j}}$ $= \{(x_i, y_i)\}_{i=1}^{n_j}$ be a subset S$_j$ of the original training set {\sc O}, where $n_j \ll N$. Let {\sc $\pi_j$} represent an input perturbation applied only to {\sc O}${\sc _{S_j}}$, producing the perturbation/challenge set {\sc $\Pi_j^{S_j}$} $= \{\pi_j(x_i), \pi_j(y_i)\}_{i=1}^{n_j}$. This results in $m$ perturbation sets $\{\Pi_j^{S_j}\}_{k=1}^{m}$, where perturbation $\pi_j$ is applied to subset $S_j$, respectively. We use {\sc P$_j$} as shorthand for the final perturbation set {\sc $\Pi_j^{S_j}$}. We evaluate the performance of model {\sc{M}} on held-out perturbation set samples {\sc Q}$_j$ for \(j = 1, \ldots, m\). Each {\sc Q}$_j$ serves as the test set specifically tailored for perturbation $\pi_j$.
\paragraph{Terminology.}  Given a pre-trained language model (PLM) denoted by {\sc M}, fine-tuned on the original (unperturbed) training set {\sc O} for a natural language processing (NLP) task {\sc T}.
\[
{\sc O} = \{(x_i, y_i)\}_{i=1}^{N}
\]
where ${(x_i, y_i)}$ represent the $i^{\text{th}}$ sample-label pair in the dataset. Let $\{\pi_j\}_{j=1}^{m}$ represent input perturbations, where $m$ is the number of distinct perturbations available.
For each perturbation $j$, let {\sc O}$_{S_j}$ be a subset $S_j$ of the original training set {\sc O}, where $n_j \ll N$:
\[
{\sc O}_{S_j} = \{(x_i, y_i)\}_{i=1}^{n_j}
\]
Perturbation $\pi_j$ is applied only to {\sc O}$_{S_j}$, producing the perturbation/challenge set {\sc $\Pi_j^{S_j}$}:
\[
{\sc \Pi_j^{S_j}} = \{ (\pi_j(x_i), \pi_j(y_i)) \}_{i=1}^{n_j}
\]
This results in $m$ perturbation sets $ \{\Pi_j^{S_j}\}_{j=1}^{m} $ where perturbation $\pi_j$ is applied to subset $S_j$. We use {\sc P}$_j$ as shorthand for the final perturbation set {\sc $\Pi_j^{S_j}$}.
We evaluate the performance of model {\sc M} on held-out perturbation set samples {\sc Q}$_j$ for $j = 1, \ldots, m$.
Each {\sc Q}$_j$ serves as the test set specifically tailored for perturbation $\pi_j$.

\subsection{Multi Model Single Set Inoculation}% ({\sc MMSS})}
\label{sec:multimodelunisetplm}
We fine-tune our PLM model using {\sc k} samples extracted from a challenge set {\sc P}$_j$. This fine-tuning across each {\sc P}$_j$ sets, results in an array of robust models each designated as {\sc RM$_{j}$}. We subsequently evaluate these models' performances across held-out challenge test sets, {\sc Q}$_j$ for every $j \in N$. This evaluation probes the efficacy of inoculating models on a singular set in enhancing—or possibly undermining—performance on test sets and different challenge/pertubation sets. While this \textit{multi model single set} framework generates multiple robust models, a clear downside emerges: as the variety of perturbation types grows, managing multiple models becomes impractical.

\subsection{Single Model Multi Set Inoculation}% ({\sc SMMS})}
\label{sec:unimodelmultisetplm}
To alleviate the complexity of managing multiple robust models, we propose cultivating a universal robust model({\sc RM}) that remains immune to various perturbations in input data. We put forth three distinct fine-tuning strategies for the same:
    
    \textbf{Sequential ({\sc Seq}):} The model is fine-tuned using {\sc k} samples from each challenge set {\sc P}$_j$ sequentially (specified by fixed {\sc Order}), resulting into a final robust model {\sc RM}.
    
    \textbf{Mixed-Training ({\sc Mix}):} In this strategy, a composite dataset, termed {\sc P$_{\sc M}$}, is fashioned by randomly selecting {\sc k} samples from all challenge sets , \{{\sc P}$_j$\}$_{j=1}^{m}$. 
    Subsequently, the model {\sc M} is fine-tuned using the aggregated {\sc P$_{\sc M}$}. In our implementation, we adopt a uniform, random sampling approach.
    
    \textbf{Dynamic Mix-Training ({\sc DynMix}):} This approach mirrors mixed-training but introduces variability in sample sizes across different challenge sets, denoted as {\sc k}$_1$, {\sc k}$_2$, and so on. Additionally, the sampling method can be unique (e.g. uniform or weighted) for each perturbation challenge set.

Given that all three finetuning outlined strategies revolve around data sampling and culminate in a singular robust model {\sc RM}, we refer this as the \textit{single model multi set} paradigm.
\subsection{Inoculation via. Prompting for LLM}
\label{sec:plminnoculation}
Fine-tuning LLMs on challenge sets is costly. In contrast, prompt tuning is quicker and more effective for many NLP tasks \citep{shin2023prompt}. Therefore, we use prompt finetuning for robustness evaluation of LLMs. 

\textbf{Original Prompt ({\sc OP}).} We design a prompt encapsulating the \textit{task} description. We also add illustrative instances (as \textit{exemplars}) from original sets ({\sc O}) which serve as main guiding posts (a.k.a few shot). Each exemplar is enriched with a rationale, mirroring a \emph{chain of thought} CoT prompting \cite{wei2023chainofthought}. This allows us to investigate the effectiveness of the perturbations $\pi_j$ on LLMs as a baseline under input perturbations. 
Here, we consider two variants of LLM prompting: 

    (a) \textbf{Zero-shot ({\sc OP$_{\text{ZS}}$})}. We create a prompt template consisting of only the description of the task, without any exemplars or reasoning chains.
    
    (b) \textbf{Few-shot with CoT ({\sc OP$_{\text{CoT}}$)}.} Here, we consider NLI task description along with few shot exemplars taken from the original set {\sc O} their reasoning chains a.k.a. CoT. 

 \textbf{Single Exemplars Multiple Prompts ({\sc SEMP})}: For each perturbation type, denoted as $\pi_j$, we construct a prompt that combines the task description, respective perturbation description, and exemplars from {\sc O} and {\sc P$_j$}. The exemplars are accompanied by corresponding labels and a reasoning chain (CoT). This results in multiple prompts, each tailored to a specific perturbation $\pi_j$. We call this approach \textit{single exemplars multiple prompts}, similar to \textit{multi model single set} (refer sec. \ref{sec:multimodelunisetplm}).

 \textbf{Multiple Exemplars Single Prompt ({\sc MESP}) :} Here, we consider descriptions and exemplars of all perturbations ($\forall \pi_j$) in a single prompt. We create a prompt by combining multiple exemplars corresponding to each perturbation $\pi_j$, sampled from {\sc P$_j$}, similar to \textit{single model multi set} in section \ref{sec:unimodelmultisetplm}. Here, the prompt contains the task description, a description of all perturbations, and exemplars from the original set {\sc O} and each of the challenge sets ({$\forall_j$ \sc P$_j$}). Given token length constraints, a trade-off between the detail of perturbation descriptions and the number of perturbation exemplars results in two variants: 
 
 (a) \textbf{Mixed Prompting Instructional ({\sc MESP$_{\text{MPI}}$}}): In this prompt, the perturbation description is emphasized while reducing the number of exemplars. 
 
 (b) \textbf{Mixed Prompting Exemplar ({\sc MESP$_{\text{MPE}}$}}): Here more perturbation exemplars are sampled and each perturbation's description is shortened.

%% file: Experiments.tex
\label{sec:case_study}
\paragraph{Original Dataset {(\sc O}).} We utilize the tabular-NLI dataset, \parentDatasetName \citep{gupta-etal-2020-infotabs}, along with its adversarial perturbations, as detailed in \citealp{shankarampeta2022enhancing}. The \parentDatasetName dataset features a wide range of table domains, categories, and keys, covering various entity types and forms. It includes three test splits: $\alpha_1$ (original test set), $\alpha_2$ (adversarial set), and $\alpha_3$ (zero-shot or out-of-domain set). 

\paragraph{Perturbed Challenge Datasets {(\sc P, Q)}.}
Our dataset incorporates perturbations from \citealp{shankarampeta2022enhancing}, enhanced using tools such as TextAttack \citep{morris-etal-2020-textattack} and NLP Checklist \citep{ribeiro-etal-2020-beyond}, alongside manual adjustments. Each perturbation specifically targets the hypothesis of an input sample. For every perturbation type, we create challenge sets of up to 1,500 samples. Only those samples that are pertinent post-perturbation are selected. When the number of such samples exceeds 1500, we narrow down to the most diverse 1500 samples using Fixed-Size Determinantal Point Processes ($\mathit{k}$-DPPs) \citep{kulesza2011k}. Perturbations used for Tabular-NLI tasks are 
Character-level perturbation (\emph{char}, C), Negation-type perturbation (\emph{neg}, N), Numeric perturbation (\emph{num}, M), Location perturbation (\emph{loc}, L) and Paraphrasing perturbation (\emph{stan}, S) (refer Figure~\ref{fig:Sample}).

\paragraph{Train/Test.} 
(a.) \emph{BERT Based Models (PLM) :} For any perturbation type, we represent {\sc Q}$_j$ consisting of 1000 examples for testing and {\sc P}$_j$ consisting of 500 examples for fine-tuning. We define the union of all challenge test sets as $\text{{\sc Q}}=\{\cup_j^m \text{{\sc Q}}_j\}$ and the training set as $\text{{\sc P}}=\{\cup_j^m \text{{\sc P}}_j\}$. 
%\tk{Q' is subset and refered as it is for LLMs}

(b.) \emph{Large Language Models (LLM) :} As LLMs inference is costly we limit our evaluations to 300 random samples from $\text{{\sc Q}}_j$, where $\text{{\sc Q}}_j$ contains original premise and perturbed hypothesis using perturbation $\pi_j$. ${\text{{\sc Q'}}_j} $ contains the original premise along with the corresponding unperturbed hypothesis as pairs. We evaluate performance on both ${\text{{\sc Q'}}_j} $ and ${\text{{\sc Q}}_j} $ to access if the LLM model forgets the original input distribution after fine-tuning on perturbation sets.

\textbf{Table Representation.} In line with \citealp{neeraja-etal-2021-incorporating}, we employed alignment techniques \citep{yadav-etal-2020-unsupervised} to eliminate distracting rows (DRR). We selected the top-8 rows for table representation as a premise (DRR@8), enhancing accuracy through evidence-based grounding of relevant information for hypothesis labeling.
%This enables a more accurate representation through evidence grounding of the labeled information relevant to the hypothesis.

 \textbf{Evaluation Metric.} We use accuracy which is equivalent to the micro-f1 score for the NLI task where the label for each example can be only one of entailment \entailment, contradiction \contra, neutral \neutral. The improvement over the multi-challenge sets is considered by taking the average of the improved performance over each challenge set $\sc Q_{j}$ %$\sc Q_{\pi_j}$ 
and this is used as the score($\mu$) for multi-perturbation setting. Implementation and hyperparameter details for all experiments are mentioned in Appendix \ref{sec:appendix:implementationdetails}.

\subsection{Fine-tuning BERT Based Model}
\label{subsec:finetunebert}
We use \robertaL \citep{liu2019roberta} as the baseline model fine-tuned on \parentDatasetName train set. This baseline model is henceforth referred to as \model. We test the baseline model on test sets from {\sc O} and {\sc Q}. By testing on {\sc Q} we attempt to demonstrate the effect of the different perturbations $\pi_C, \pi_N, \pi_M, \pi_L, \pi_S$ on \model. 

\textbf{Multi Model Single Set Inoculation.} \model is further fine-tuned on different types of challenge sets($\text{\sc P}_{j}$), resulting in multiple robust models. 

\textbf{Single Model Multi Set Inoculation.} We propose three different strategies:

\begin{itemize}
    \item \textbf{Sequential ({\sc Seq}):} We perform sequential fine-tuning of \model across various challenge sets. The training order ({\sc Order}) for fine-tuning is based on average baseline model performance across challenge sets. Sequential fine-tuning often leads to catastrophic forgetting of previously learned perturbations \citep{kirkpatrick2017overcoming,goodfellow2013empirical}. To mitigate this, we propose two alternative strategies designed to minimize this effect. %Our sequencing strategy aims to minimize the potential for catastrophic forgetting \tk{this isnt correct, cat for is handled by mix and dynmix} \citep{kirkpatrick2017overcoming,goodfellow2013empirical} induced by subsequent fine-tuning on challenge sets.
    \item \textbf{Mixed-Training ({\sc Mix}):} Here, the \model is fine-tuned samples obtained by mixing {\sc k} instances drawn from each of the challenge sets $\text{{\sc P}}_M, \text{{\sc P}}_N, \text{{\sc P}}_L, \text{{\sc P}}_C, \text{{\sc P}}_S$. Here, {\sc k} is an hyper-parameter, set equal to $500$ examples, as discussed in section 3.1.
    \item \textbf{Dynamic Mix-Training ({\sc DynMix}):} This is similar to {\sc Mix}, except the number of samples drawn from each of the challenge sets is different.  The number of samples is determined by the inverse of the baseline (higher baseline metrics results in lower number of samples) accuracy for \model for challenge sets {\sc P}$_j$.
\end{itemize}

\subsection{LLM Prompting}\label{LLM:promptEngineering}
We used GPT-3.5 with low temperature of $0.3$, LLaMA-2 after quantization using QLoRA \citep{dettmers2023qlora}, Mistral \cite{jiang2023mistral} and Flan-T5 series \cite{chung2024scaling}.%\tk{We can rephrase this line, as we are also reporting for other models along with gpt3.5:} 
We develop methodologies for LLMs that rely solely on prompting and exclude fine-tuning (except for GPT-3.5 where we also report fine-tuning results). The LLM prompt design for our experiments, is detailed in Table~\ref{prompt-template}, comprises five sections, with demonstration section being optional. 
\begin{table}[!tbh]
\renewcommand{\arraystretch}{1.1}
\centering
\small
\setlength{\tabcolsep}{3pt} % Adjust the spacing between columns
\scalebox{0.78}{
\begin{tabular}{@{}p{2.0cm}|p{7.4cm}@{}}
\hline
\multicolumn{2}{c}{\bf Broad Prompt Template}\\
% Broad Template \\
\hline
\textbf{{NLI Task Explanation}} & In this task, we will ask you to make an inference about the information presented as the premise. We will show you a premise and a hypothesis... \\
\hline
\textbf{{Perturbation Awareness}} & The concept of numeric and character typos in questions is important for maintaining the integrity and meaning of a sentence... \\
\hline
\textbf{{Description of Limitation}} & It is very important and critical that you do not use information other than the premise that you may know if you believe that it is not generally known... \\
\hline
\textbf{{Answering}} & (Restriction for Answering) Answer with an explanation in the following format, restricting the answer to only one of the following: "yes" or "no" or "it is not possible to tell" + <Answering Format> \\
\hline
\textbf{{Demonstrations}} & Demonstrations from different sets with reasoning (CoT). \\
\hline
\end{tabular}}
\vspace{-0.5em}
\caption{Prompt Structure used in LLMs}
\label{prompt-template}
\vspace{-1.0em}
\end{table}

\textbf{Original Prompt ({\sc OP}).} This is the original prompt zero shot ({\sc OP$_{\text{ZS}}$}) setting with NLI task description. In CoT setting ({\sc OP$_{\text{CoT}}$}), we define our few shot setting, where exemplars are sampled from original training dataset O. 

\textbf{Single Exemplars Multiple Prompts ({\sc SEMP}).} For a designated perturbation \( \pi_j \) from the set \( \{\pi_C, \pi_N, \pi_M, \pi_L, \pi_S\} \), our prompts integrate the NLI task outline, a brief on the perturbation \( \pi_j \), and its Chain of Thought (CoT) exemplars sourced from the respective challenge set \( \text{{\sc P}}_j \). 

\textbf{Multiple Exemplars Single Prompt ({\sc MESP}).} These prompts contain NLI task description, description of all perturbations $\pi_j \in \{\pi_C, \pi_N, \pi_M, \pi_L, \pi_S\}$ and exemplars sampled from each challenge set {\sc P}$_j$ $\in$ $\{\text{{\sc P}}_M, \text{{\sc P}}_N, \text{{\sc P}}_L, \text{{\sc P}}_C, \text{{\sc P}}_S$\}. Here , we consider two different prompts settings {\sc MESP$_{\text{MPI}}$} and {\sc MESP$_{\text{MPE}}$}, as described earlier in section \ref{sec:plminnoculation}.

Complete prompt examples for each case can be found in Appendix \ref{sec:appendix:implementationdetails}.
% \vspace{-0.75em}

%% file: Results.tex
% \vspace{-0.75em}

Our experiments answer the following questions:-
%\tk{Can we make some stuff bold here or is it okay}
 \begin{itemize}
     \item Do input perturbations pose a challenge for Language Models (PLMs and LLMs)? 
     \item How does the approach of single model fine-tuning on multiple perturbation sets compare to multiple models fine-tuning on a single perturbation set in terms of inoculation?
    \item Do details perturbation descriptions, multiple exemplars, and Chain of Thought (CoT) prompts enhance LLM robustness? 
    \item What holds greater importance for LLM prompting: the quality of descriptions or the quantity of exemplars?
 \end{itemize}

\subsection{Results: Bert Style Models (PLM)}\label{sec:results_roberta}
\paragraph{Multi Model Single Set Inoculation.}
The baseline performance of \model original and challenge sets is shown in Table~\ref{table-cross-testing-roberta-large}. We also report the performance after fine-tuning each challenge set in the same table.

\begin{table}[!ht]
\renewcommand{\arraystretch}{1.1}
    \centering
    \small
    \setlength{\tabcolsep}{2.75pt} % Adjust the spacing between columns
    \scalebox{0.85}{
        \begin{tabular}{@{}l|ccc|ccccc}
            \hline
            \multicolumn{1}{c}{} & \multicolumn{3}{c}{\textbf{Original Test Sets}} & \multicolumn{5}{c}{\textbf{Challenge Test Sets}} \\
            \hline
            \textbf{Train/\ Test} & $\boldsymbol{\alpha_1}$ & $\boldsymbol{\alpha_2}$ & $\boldsymbol{\alpha_3}$ & \textbf{char} & \textbf{neg} & \textbf{num} & \textbf{loc} & \textbf{stan} \\
            \hline
            baseline &72.72 & \bf 64.83 & 62.33 & 57.30 & 46.90 & 67.20 & 70.20 & 67.10 \\
            \hline
            char & \bf 75.28 & 63.83 & \bf 63.33 & \textbf{59.20} & 43.70 & 64.30 & 66.00 & \bf 68.30 \\
            neg & 66.94 & 64.56 & 58.06 & 52.80 & \textbf{71.90} & 69.60 & 69.70 & 62.40 \\
            num & 62.06 & 60.83 & 52.50 & 47.30 & 49.60 & \textbf{85.40} & 83.00 & 57.60 \\
            loc & 55.78 & 58.67 & 49.67 & 47.40 & 53.90 & 84.60 & \textbf{86.10} & 53.50 \\
            stan & 73.56 & 62.61 & 60.44 & 58.30 & 40.80 & 70.30 & 67.80 & 66.80 \\
            \hline
        \end{tabular}
    }
\vspace{-0.5em}
\caption{\textbf{Multi-model Uniset Inoculation:} \model when fine-tuned on one of the challenge sets ({\sc P}$_j$), but tested on all challenge sets ({\sc Q}$_j$) with number of sample used equal 500. }
 \vspace{-1.0em}
    \label{table-cross-testing-roberta-large}
\end{table}

\textit{Analysis.} 
\begin{inparaenum}[(a.)]
    \item Baseline performance of \model on challenge sets is notably lower than on original sets, emphasizing PLMs' vulnerability to input perturbations.
    \item Fine-tuning via single-set inoculation significantly bolsters the model against specific perturbations, improving negation accuracy by +25 points from baseline.
    \item Despite fine-tuning, the model's robustness to paraphrasing remains largely unchanged.
    \item While the fine-tuned model excels against specific perturbations, it struggles with others. Interestingly, character perturbations inadvertently boost its proficiency in challenges like paraphrasing.
    \item Inoculation effects vary: character set inoculation enhances performance on original test sets, while number and location decrease performance in both original and challenge test sets.
\end{inparaenum}

\paragraph{Single Model Multi Set Inoculation.} We present results on Sequential training ({\sc Seq}), Mixed Training ({\sc Mix}), and Dynamic Mixed Training ({\sc DynMix}) in Table \ref{table-all-fine-tuning}. 
\begin{table*}[!htb]
% \vspace{-0.5em}
\renewcommand{\arraystretch}{1.1}
\centering
\small
\setlength{\tabcolsep}{8pt} % Adjust the spacing between columns
\scalebox{1.0}{
\begin{tabular}{p{0.25cm}|c|ccc|ccccc|c@{}}
%\hline
& \multicolumn{1}{c}{} & \multicolumn{3}{|c}{\textbf{Original Sets}} & \multicolumn{5}{|c}{\textbf{Challenge Sets}} & \multicolumn{1}{|c}{} \\
\hline
& $\boldsymbol{K}$/SEQ-Type & $\boldsymbol{\alpha_1}$ & $\boldsymbol{\alpha_2}$ & $\boldsymbol{\alpha_3}$ & \textbf{char} & \textbf{neg} & \textbf{num} & \textbf{loc} & \textbf{stan} & $\boldsymbol{\mu}$ \\
\hline
& baseline &  72.72 & 64.83 & \bf 62.33 &  57.30 & 46.90 & 67.20 & 70.20 &  67.10 & - \\
\hline
 \multirow{4}{=}{\begin{sideways}\textbf{{\sc SEQ}}\end{sideways}}
& \emph{COL-ASC} & 61.67 & 60.94 & 50.11 & 48.80 & 54.60 & \bf 85.40 & 85.40 & 56.60 &  4.42 \\
& \emph{COL-DSC} & \bf 74.67 & 62.72 & 60.44 & \bf 58.90 & 57.30 & 56.10 & 65.30 & \bf 68.00 & -0.62 \\
& \emph{ROW-ASC} & 55.00 & 58.11 & 47.22 & 46.80 & 50.90 & 84.50 & \bf 85.90 & 51.30 & 2.14\\
& \emph{ROW-DSC} & 73.44 & 63.39 & 57.44 & 56.50 & 45.10 & 60.00 & 71.60 & 65.80 & -1.94\\
\hline
 \multirow{5}{=}{\begin{sideways}\textbf{{\sc Mix}}\end{sideways}}
& 100 & 70.40 & \bf 65.16 & 59.48 & 56.00 & 58.48 & 78.78 & 78.50 & 66.04 & 5.82 \\
& 200 & 70.42 & 65.06 & 59.21 & 56.86 & 59.50 & 80.94 & 80.36 & 64.68 & 6.73 \\
& 300 & 71.92 & 64.54 & 59.49 & 56.50 & 61.30 & 81.22 & 79.68 & 65.12 & 7.02 \\
& 400 & 72.11 & 64.48 & 59.78 & 56.58 & 63.70 & 81.60 & 80.38 & 64.64 & 7.64 \\
& 500 & 72.62 & 64.34 & 59.20 & 56.98 & \bf 66.06 &  82.02 & 80.52 & 65.64 & \textbf{8.50} \\
\hline
\multirow{3}{=}{\begin{sideways}{\sc DynMix}\end{sideways}}
& 500 & 71.28 & 64.42 & 60.39 & 56.26 & 59.22 & 77.84 & 76.24 & 65.38 & 5.25 \\
& 1000 & 71.07 & 64.72 & 59.60 & 57.04 & 63.24 & 79.94 & 79.06 & 65.50 & 7.22 \\
& 1500 & 72.07 & 64.81 & 59.73 & 56.50 & 65.42 &  80.84 & 79.54 & 65.64 & 7.85 \\
%\hline
\end{tabular}}
\caption{\textbf{ Single Model Multi Set Fine tuning Strategies Results:} For \textsc{Seq} Results , \model is Sequential Trained with 500 samples from each {\sc P$_j$}. Here, \textbf{{\sc col-asc: csnlm}}, \textbf{{\sc col-dsc: mlnsc}}, \textbf{{\sc row-asc: scnml}}, \textbf{{\sc row-dsc: lmncs}} are the sequence types and $\mu$ is the average improvement. For {\sc Mix} Results, \model fine-tuned on $\boldsymbol{K}$ equal samples from different perturbation sets {\sc P}$_j$. For {\sc DynMix} Results, \model fine-tuned on total of $\boldsymbol{K}$ samples taken from {\sc P$_j$} in ratios mentioned in the \sc{DynMix} section below.}
\label{table-all-fine-tuning}
\vspace{-1.0em}
\end{table*}

\textbf{\textsc{Seq}.} Table~\ref{table-all-fine-tuning} presents the results using Sequential Training ({\sc Seq}). The method trains \model on varied challenge sets in distinct sequences. For instance, {\sc Order} {\sc MNLCS} with {\sc k} samples implies training sequentially on subsets of $\{\text{{\sc P}}_M, \text{{\sc P}}_N, \text{{\sc P}}_L, \text{{\sc P}}_C, \text{{\sc P}}_S$\} of size {\sc k}. This is denoted as \textsc{Seq$_{\text{MNLCS}}$}.

\textit{Terminology.} To define the sequence we consider 

(a.) \textit{Column Wise Average.} This configuration assesses the aggregate impact of fine-tuning across all perturbations on each individual perturbation. 

(b.) \textit{Row Wise Average.} This configuration evaluates the aggregate impact of fine-tuning on an individual perturbation against all other perturbations. For more details on the metrics refer to Appendix \ref{sec:appendix:implementationdetails}.

We compute both COL and ROW values for each perturbation. By sorting these values, we derive sequences in ascending and descending order, yielding the {\sc col-asc, col-dsc, row-asc, row-dsc} as the {\sc order} sequences. 

\textit{Analysis.} Sequential training introduces the forgetting issue \citep{he-etal-2021-analyzing,chen-etal-2020-recall}, where models forget sets trained on earlier in the sequence. (a.) With column-wise averages, we capture how easy a perturbation $\pi_j$ is to learn by fine-tuning on other perturbations by testing improvement in accuracy on set {\sc Q$_j$}. Therefore in the {\sc order col-asc}, an "easier" perturbation appears later and hence improves the average performance. (b.) With row-wise averages, we capture how much fine-tuning on {\sc P$_j$} improves the overall performance of other perturbation types. Hence, in the {\sc order row-asc} with samples from {\sc P$_j$} wherein $\pi_j$ has a higher score appearing later, benefit other better perturbation effectively. 

\textbf{\textsc{Mix}.}
Table~\ref{table-all-fine-tuning} presents the outcomes from multi-set inoculation using mixed training.

%\tk{We can re-refer to cat. forgetting here somewhere}
\textit{Analysis.} Models trained via mixed training outperform those from {\sc SEQ}. As we increase the number of samples for fine-tuning, we notice consistent gains across most challenge sets and original test sets. The most prominent improvements are seen in the negation and location sets. While there's a minor performance dip in some original and challenge sets, it's less pronounced compared to results from single-set inoculation and {\sc SEQ}.

\textbf{\textsc{DynMix}.} Table~\ref{table-all-fine-tuning} displays the results from dynamic mixed training. The sample ratio of $0.223:0.278:0.171:0.156:0.172$ for $C:N:M:L:S$ was determined based on the inverse of baseline performance values (i.e., poorer baseline performance warrants more samples from that perturbation set).

\textit{Analysis.} Though the dynamic mixed training surpasses {\sc SEQ}, it only edges out the mixed training approach when utilizing a total of 1000 and 1500 samples for fine-tuning for {\sc k} = 200, 300. This shows that dynamically altering challenge set size improves single model multi-set inoculation. \emph{In conclusion, multi-set inoculation produces robust models than single-set. Further, the {\sc Mix} and {\sc DynMix} strategies for fine-tuning stand out as more resilient compared to {\sc Seq}}.

\paragraph{Ablation Experiments.} 

(a) \emph{Fine tuning on a subset of Perturbations.} Above {\sc Mix} and {\sc DynMix} require access to all perturbations during fine-tuning, which increases dataset and computation costs. To assess whether robust models can be obtained via fine-tuning on a subset of perturbation sets, we ran experiments using a subset of perturbations. The results are shown in Appendix \ref{sec:appendix:additional_results}. Our results show that although there are performance improvements while fine-tuning on subsets of perturbation. Nevertheless, the optimal subset of available perturbations for the task remains elusive and cannot be found empirically.

(b) \emph{Results on Out of Distribution Perturbations.} Assessing the model's performance against unseen perturbations is vital for robustness. Such evaluation reveals the model's ability to adapt to new and unexpected changes. We created approximately 100 samples (with nearly equal numbers of E, C, N labels) of a new \textbf{\sc{Word-Swap}} perturbation type. The results are shown in Appendix \ref{sec:appendix:additional_results}. We observe fine-tuning with more samples using the {\sc Mix} strategy enhances model robustness against unseen perturbations, further validating our approach.

\input{prompt-enginerring-results}

% \vspace{-0.85em}

%% file: prompt-enginerring-results.tex
% \vspace{-1.0em}
\subsection{Results: Large Language Models (LLMs)}

\textbf{Original Prompt.}  Table~\ref{llm-op-zs-cot} shows the results for {\sc OP$_{\text{ZS}}$} and {\sc OP$_{\text{CoT}}$}, respectively. Results on other open source models in Appendix \ref{sec:appendix_additional_results}.

\input{tables/table-op-cot-zs}

\textit{Analysis.} \begin{inparaenum}[(a.)] On the Original Zero-Shot Prompts we observe that, 
    \item Comparing the results of challenge datasets {\sc Q$_j$} and their unperturbed version sets {\sc Q'$_j$} reveals that LLMs similar to PLMs are also sensitive to input data perturbations. 
    \item However, the Flan-T5 series, specifically XL and XXL, performs significantly better than other LLMs as it's fine-tuned specifically for the NLI task \citep{chung2022scaling}. Even the drop in performance due to data perturbation is relatively less.
    \item The poor performance of relatively smaller LLMs, such as LLaMA-2-13b, demonstrates the ineffectiveness of such models in responding to an instruction prompt. 
    \item One reason for lower performance on original numerical set ({\sc Q'$_M$)}, is due to model inability to handle mathematical reasoning \citep{wallace2019nlp,min2021recent,hendrycks2021measuring,imani-etal-2023-mathprompter}. 
    Additionally, we find that all models enhanced with CoT (Table \ref{llm-op-zs-cot}) outperform those using Zero Shot original prompts. This suggests that simply adding exemplars can enhance a model's resilience to perturbations. 
\end{inparaenum}

 % \noindent 
 \textbf{Single Exemplars Multiple Prompts ({\sc SEMP})}: Table~\ref{table-single-aware-a} presents results for GPT-3.5, with diagonal elements as an analog to single set inoculation. LLaMA-2 results are in Table~\ref{table-llama-2-single-aware-b}.%\tk{We have to specify that LLaMA results are restricted to just the diagnoal, or remove them altogether The tables don't go together}\vl{ref}
 
\input{tables/table-single-aware}
\textit{Analysis.} From Tables~\ref{table-single-aware-a} and~\ref{table-llama-2-single-aware-b}, it's evident that incorporating an input perturbation explanation within the prompt enhances the model's accuracy. The results in Table ~\ref{table-single-aware-a} suggest that even a singular perturbation explanation prompts the model to anticipate other perturbations, essentially priming it for a noisy environment. This adaptability is especially pronounced for character perturbations, where improvements span across all challenge sets. Comparisons with instructional prompts and few-shot results show that demonstrations with perturbation explanations improve performance.

\begin{figure}
    \centering
    \includegraphics[width=1.0\linewidth]{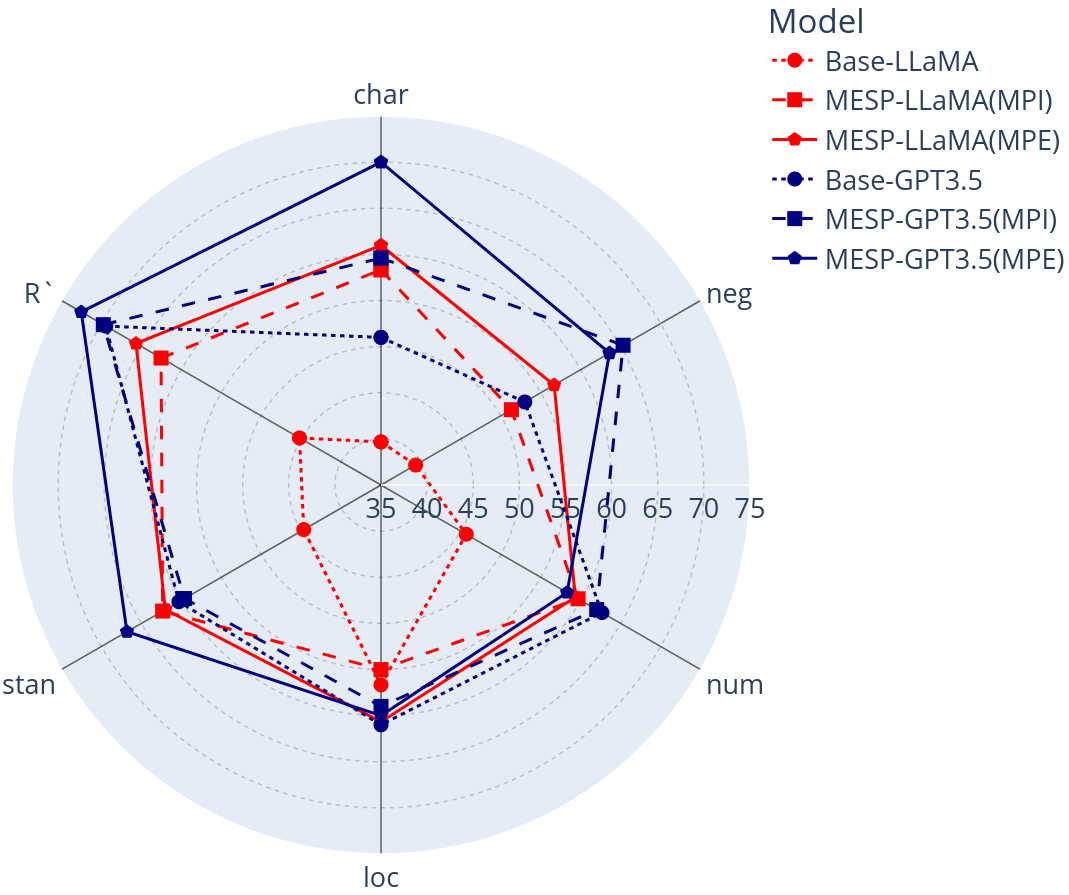}
    \caption{\textbf{MESP Results on LLaMA-2-13b and GPT-3.5}.\textit{LLaMA-2 refers to LLaMA-2-13b.}}
    \label{table-mix-llms-gpt-llama}
    \vspace{-1.0em}
\end{figure}
\textbf{Multiple Exemplars Single Prompts ({\sc MESP})}: The results for \textbf{MPI} and \textbf{MPE} are in Figure ~\ref{table-mix-llms-gpt-llama}.

\textit{Analysis.} Both models show marked improvement with mixed prompting, indicating that LLMs, when guided with perturbation descriptions and examples, yield more stable outputs. The superior performance of MPE over MPI suggests that including more examples in prompts is more beneficial than detailed perturbation descriptions.

In conclusion, LLMs too face challenges with input perturbations. Simply explaining one perturbation primes the LLM to consider others.  Our findings show that a mixed prompting approach with several perturbation instances and brief explanations improves robustness.

\begin{figure}
    \centering
    \includegraphics[width=1.0\linewidth]{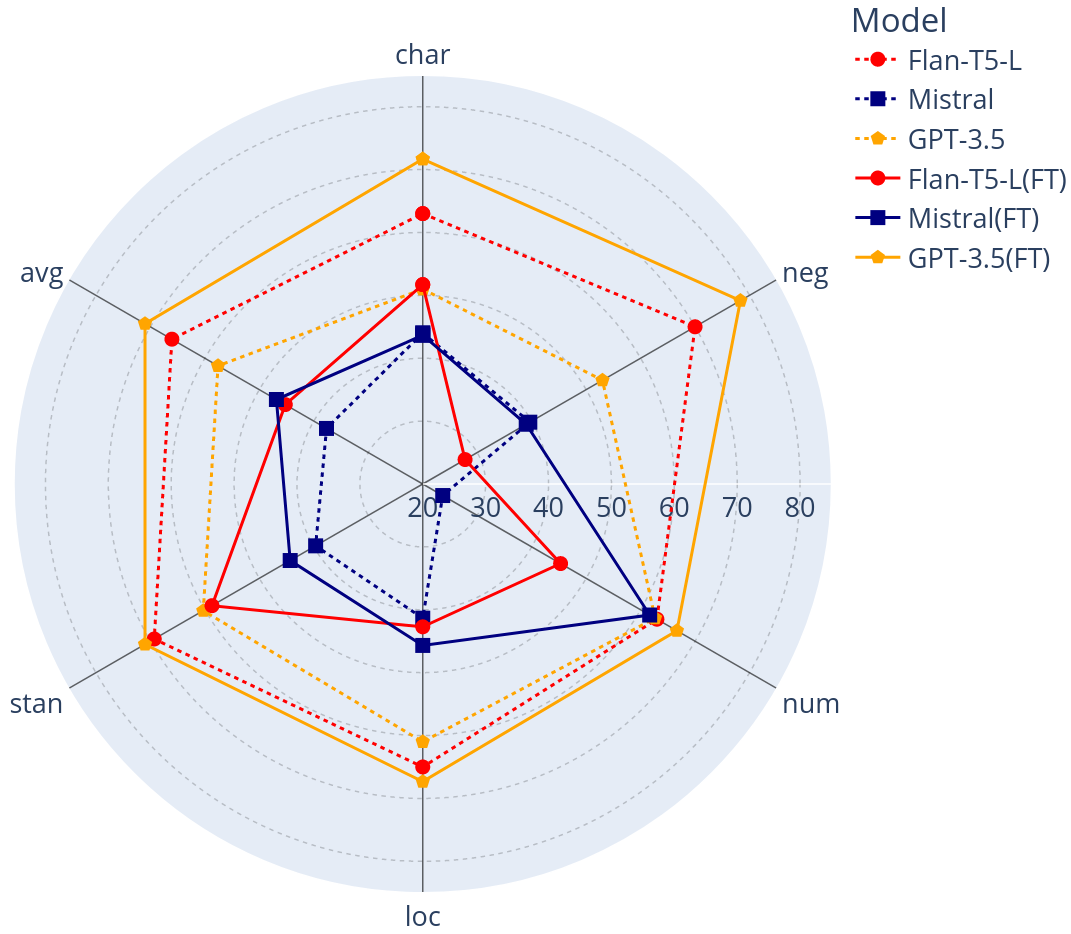}
    \caption{Fine tuning results for Flan-T5-L-0.8b, Mistral-7b-instruct-v0.2 and GPT-3.5-turbo on perturbed sets and average of performance.
\textit{FT refers to Fine-Tuning results and w/o FT refers to {\sc OP$_{\text{ZS}}$} results.}}
    \label{fig:table_fine_tuning}
    \vspace{-1.0em}
\end{figure}

\textbf{Fine-tuning on LLMs.}
\label{fine_tune_sec}
While our work primarily focuses on in-context learning for LLMs, we also examine the effects of fine-tuning LLMs on perturbation sets, results shown in Figure \ref{fig:table_fine_tuning}. %Table ~\ref{tab:fine_tuning_LLM}. 
We can see that for Mistral and GPT-3.5 the fine tuning with the perturbation set using the mix training approach increases the models' performance. Whereas for the Flan-T5-L model the fine tuning does not improve the model's performance. 

%% file: tables/table-op-cot-zs.tex
\begin{table}[!ht]
\vspace{-0.25em}
    \centering
    \renewcommand{\arraystretch}{1.1}
    \small
    \setlength{\tabcolsep}{4.6pt} % Adjust the spacing between columns
    \scalebox{0.80}{
        \begin{tabular}{p{0.25cm}|p{0.25cm}|l|cccccc}
            % &&  \multicolumn{6}{c}{\bf Metric Scores} \\
            % \hline
            &&\textbf{Model} & \textbf{char} & \textbf{neg} & \textbf{num} & \textbf{loc} & \textbf{stan} & \textbf{avg.} \\
            \hline
            \multirow{6}{=}{\begin{sideways}\textbf{{\sc OP$_{\text{ZS}}$}}\end{sideways}}
            &\multirow{3}{=}{\begin{sideways}\textbf{Q'}\end{sideways}}
            &Flan-t5-XXL & \textbf{70.60} & \textbf{77.30} & \textbf{69.00} & \textbf{74.00} & \textbf{79.00} & \textbf{73.98} \\
            &&LLaMA-2-70b & 59.00 & 63.60 & 64.60 & 67.00 & 60.00 & 62.84 \\
            &&GPT-3.5 & 68.00 & 69.00 & 68.66 & 71.60 & 70.00 & 69.45 \\
            \cline{2-9}
            &\multirow{3}{=}{\begin{sideways}\textbf{Q}\end{sideways}}
            &Flan-t5-XXL & 63.00 & 70.00 & 63.00 & 65.00 & 69.30 & 66.06 \\
            &&LLaMA-2-70b & 54.00 & 51.60 & 49.60 & 57.00 & 54.30 & 53.30 \\
            &&GPT-3.5 & 51.00 & 53.00 & 62.66 & 61.00 & 60.30 & 57.59 \\
            \hline
            \multirow{6}{=}{\begin{sideways}\textbf{{\sc OP$_{\text{CoT}}$}}\end{sideways}}
            &\multirow{3}{=}{\begin{sideways}\textbf{Q'}\end{sideways}}
            &LLaMA-2-13b  & 63.67 & 69.33 & 66.33 & 61.00 & 61.00 & 64.27 \\
            &&LLaMA-2-70b & 68.6 & 72.3 & 76.3 & 67.3 & 69.6 & 70.82 \\ 
            &&GPT-3.5 & 68.30 & 76.30 & 68.00 & 73.00 & 75.30 & 72.18 \\ 
            \cline{2-9}
            &\multirow{3}{=}{\begin{sideways}\textbf{Q}\end{sideways}}
            &LLaMA-2-13b  & 61.33 & 57.00 & 57.67 & 59.33 & 60.00 & 59.07 \\
            &&LLaMA-2-70b & \textbf{63.00} & 60.00 & \textbf{63.00} & \textbf{61.30} & 66.00 & 62.66 \\ 
            &&GPT-3.5 & 63.00 & \textbf{69.60} & 59.30 & 61.00 & \textbf{68.00} & 64.18 \\ 
        \end{tabular}
    }
    \caption{(a) \textbf{Zero Shot ({\sc OP$_{\text{ZS}}$}):} Baseline Accuracies on original and perturbed sets for prompts in zero-shot setting. (b) \textbf{Few-shot with CoT ({\sc OP$_{\text{CoT}}$}):} Results using CoT prompting with exemplars sampled from {\sc O}.}
    \label{llm-op-zs-cot}
     \vspace{-1.0em}
\end{table}

%% file: tables/table-single-aware.tex
% \begin{table}[!ht]
% \vspace{-0.25em}
% \renewcommand{\arraystretch}{1.1}
% \centering
%     \small
%     \setlength{\tabcolsep}{3pt} % Adjust the spacing between columns
%     \scalebox{0.85}{
% \begin{tabular}{l|cccccc}
% \hline
% %\diagbox[innerwidth=1.9cm, height=1cm]{\textbf{Prompt}}{\textbf{Test}} & \textbf{char} & \textbf{neg} & \textbf{num} & \textbf{loc} & \textbf{stan} & \textbf{$R'$} \\
% \textbf{Prompt/\ Test} & \textbf{char} & \textbf{neg} & \textbf{num} & \textbf{loc} & \textbf{stan} & \textbf{$R'$} \\
% \hline
% baseline & 51.00 & 53.00 & 62.66 & 61.00 & 60.30 & 69.05 \\
% \hline
% char & \bf 67.60 & 65.30 & \bf 66.00 & \bf 69.00 & \bf 67.60 & 68.05 \\
% neg & 60.30 & 64.60 & 58.00 & 59.60 & 63.30 & 71.62 \\
% num & 62.30 & 66.30 & 61.00 & 60.60 & 64.30 & 70.24 \\
% loc & 62.60 & 63.60 & 61.00 & 59.30 & 64.00 & 71.30 \\
% stan & 59.00 & \bf 67.60 & 61.30 & 61.00 & 67.30 & \bf 73.76 \\
% \hline
% \end{tabular}}
% \vspace{-0.5em}
% \caption{\textbf{{\sc \textbf{SEMP}} Results on GPT-3.5:} The last column is the average performance on all sets of {\sc R$'$}.}
% \label{table-single-aware}
% \vspace{-1.25em}
% \end{table}

\begin{table}[!ht]
\centering
\begin{subtable}[b]{0.475\textwidth} % Adjust widths as needed
\renewcommand{\arraystretch}{1.1}
\centering
  \small
  \setlength{\tabcolsep}{5.75pt} 
  \scalebox{0.90}{
\begin{tabular}{l|cccccc}
% \hline
\textbf{Pr/\ Test} & \textbf{char} & \textbf{neg} & \textbf{num} & \textbf{loc} & \textbf{stan} & \textbf{$Q'$} \\
\hline
baseline & 51.00 & 53.00 & 62.66 & 61.00 & 60.30 & 69.05 \\
\hline
char & \bf 67.60 & 65.30 & \bf 66.00 & \bf 69.00 & \bf 67.60 & 68.05 \\
neg & 60.30 & 64.60 & 58.00 & 59.60 & 63.30 & 71.62 \\
num & 62.30 & 66.30 & 61.00 & 60.60 & 64.30 & 70.24 \\
loc & 62.60 & 63.60 & 61.00 & 59.30 & 64.00 & 71.30 \\
stan & 59.00 & \bf 67.60 & 61.30 & 61.00 & 67.30 & \bf 73.76 \\
% \hline
\end{tabular}}
\caption{\textbf{{\sc SEMP} Results on GPT-3.5}} 
\label{table-single-aware-a}
\end{subtable}
\hspace{2.5em}
%\hfill 
\begin{subtable}[b]{0.43\textwidth} % Adjust widths as needed
% \vspace{-0.25em}
\centering
\small
\setlength{\tabcolsep}{4.3pt} 
\scalebox{0.94}{
\begin{tabular}{p{0.78cm}|c|ccccc}
% \hline
Type &\textbf{{$\pi_j$}} & \textbf{char} & \textbf{neg} & \textbf{num} & \textbf{loc} & \textbf{stan} \\
\hline
\multirow{2}{=}{\textbf{BASE}} 
&{\sc Q}$'_j$ & 59.00 & 63.60 & 64.60 & 67.00 & 60.00\\
&{\sc Q}$_j$ & 54.00 & 51.60 & 49.60 & 57.00 & 54.30\\ 
\hline
\multirow{2}{=}{\textbf{SEMP}} 
&{\sc Q}$'_j$ & 69.00 & 71.00 & 72.00 & 72.30 & 68.60 \\ 
&{\sc Q}$_j$ & 53.00 & 58.00 & 62.00 & 62.00 & 68.30 \\ 
% \hline 
\end{tabular}}
\caption{\textbf{{\sc SEMP} Results on LLaMA-2-70b}}
\label{table-llama-2-single-aware-b}
\end{subtable}
 \vspace{-0.5em}
\caption{\textbf{{\sc SEMP} Results}: (a) The last column is the average performance on all sets of {\sc Q$'$} (b) Self-testing on perturbation $\pi_j$ with prompt for $\pi_j$ and test on {\sc{Q$_j$}} and {\sc{Q$'_j$}}.} 
\label{fig:combined_tables} 
\vspace{-1.0em}
\end{table}

%% file: related-works.tex
\paragraph{Model Robustness Issues.} Deep learning models in vision and language domains have exhibited sensitivity to adversarial examples and input distribution shifts, as highlighted in prior studies \cite{mahmood2021robustness,elsayed2018adversarial,chang2021robustness,ren2019generating,mccoy-etal-2019-right,wang2021infobert,gupta2023whispers,zheng2023noisy,zhu2023promptbench}. 
The search for model robustness in language processing has led to work on
contrast sets \citep{li2020linguistically}, Checklist \citep{ribeiro-etal-2020-beyond}, and attack algorithms \citep{li2020bert,li2018textbugger}. Ensuring model robustness is crucial \citep{wang2021measure,wang-etal-2020-cat}, as minor input changes can significantly impact performance due to model complexity and distribution overfitting \citep{glockner-etal-2018-breaking,rice2020overfitting,zhu-rao-2023-exploring,moradi2021evaluating}. Recently, \citet{zhu2023promptbench} introduce adversarial prompts to analyse model robustness to perturbation in prompts. Our work focuses on analyzing model performance with clean prompts across several perturbations/attacks on input samples simultaneously.
\paragraph{Improving Model Robustness.} 
Utilizing adversarial examples during training provides a degree of mitigation to input sensitivity of a deep learning model \citep{tong2022robust,liu-etal-2019-inoculation,yuan-etal-2023-hype,kotha2023understanding,liu2023towards}, however, it falls short of a comprehensive solution for achieving widespread robustness, as it deals only with one facet, i.e., single-set inoculation.  Our proposed framework is adept at evaluating model robustness across multiple challenge sets. Our research complements and extends the work on robustness explored in \cite{liu2023towards, lu2022multi, zheng2023noisy}. 
While \citealp{liu2023towards} integrates consistency loss and data augmentation during training, our framework applies to models already in use or deployed. Similarly, \citealp{lu2022multi} addresses dataset artifacts in natural language inference (NLI) with a multi-scale data augmentation method. In contrast, our work focuses on limited fine-tuning of pre-trained models and expands to additional dimensions of robustness. Meanwhile, \citealp{zheng2023noisy} examines LLM robustness to perturbed inputs by increasing noisy exemplars. Our study offers a broader framework for assessing the robustness of both PLMs and LLMs, using fine-tuning, improving instruction quality, and enhancing exemplars in both diversity and quantity.

%% file: Appendix.tex
\label{sec:appendix}

\subsection{Additional Results} \label{sec:appendix:additional_results}

\subsubsection{PLM results on Perturbation Subsets}

Fine-tuning on the entire set of possible perturbations necessitates access to all possible perturbations, which is infeasible. Moreover, it would demand substantial computational resources to fine-tune a robust model using strategies like {\sc Mix} or {\sc DynMix}. However, we see that there is a positive correlation between char/stan and num/loc perturbations and negative correlation between neg and other perturbations as shown in Table \ref{table-cross-testing-roberta-large}. To reduce computational and annotation costs, fine-tuning the model on a subset of perturbations can enhance overall performance across all perturbations.

Using performance correlation analysis from Table \ref{table-cross-testing-roberta-large}, we create two training subsets (a) (neg, num, loc) type perturbations (Table ~\ref{tab:in-out-1}) and (b) (char, num) type perturbations (Table ~\ref{tab:in-out-2}).(a) We selected 'char' and 'num' due to their positive correlation, which also positively impacts other perturbation sets. (b) For 'neg', 'num', and 'loc', we chose 'neg' because it's negatively correlated with all other sets, while 'loc' and 'num' are positively correlated with 'char' and 'stan'. With this set, we aimed to analyze the impact of negatively correlated sets in fine-tuning.

From Table~\ref{tab:in-out-1}, the bias detected in the mean score reveals a complex picture: as the overall mean score rises, we see an improvement in performance on perturbation types targeted during fine-tuning. However, this is contrasted by a simultaneous decrease in performance on other perturbation types. This pattern emphasizes the exclusivity of these specific perturbations and clearly illustrates the presence of a negative correlation.

From Table~\ref{tab:in-out-2} we notice that training both num and char together is not improving char perturbation accuracy. We don’t see improvement in paraphrasing as well but we don’t see a consistent decrease well (likely because num type perturbation dominates during fine-tuning process). From the above analysis it can be observed that predicting behaviour on smaller perturbation subsets is potentially complex.

\emph{Conclusion}: These further experiments underscore the importance of selecting appropriate perturbation sets for training. By applying single set cross-testing, as shown in Table~\ref{table-cross-testing-roberta-large}, we can identify sets that are positively and negatively correlated. An effective approach could be to train on negatively correlated sets and sample from positively correlated ones, which helps in reducing the total number of sets needed, without sacrificing on performance (i.e. maintaining similar performance). However, it's important to note that this selection strategy may initially demand significant computational resources. This initial computational cost stems from the need to establish performance correlations between perturbation sets, as referenced in Table \ref{table-cross-testing-roberta-large}. \input{tables/in-out-merge}

\subsubsection{PLM results on Out of Distribution Perturbation}
\textsc{\textbf{Mix$_{OOD}$}}
Assessing the model's performance against unseen perturbations is vital for robustness. Such evaluation reveals the model's ability to adapt to new and unexpected changes. We created approximately 100 samples (with nearly equal numbers of E, C, N labels) of a new \textbf{\sc{Word-Swap}} perturbation type. This involves selecting words for replacement with others, as illustrated in the example below:

\noindent \exampleParagraph{}{

\noindent\textbf{Original Hypothesis:} Josh Groban was born inside of the US.

\textbf{Perturbed Hypothesis:} Josh Groban was inside born of the US.

}

\noindent Our word-swap perturbation generation prioritizes swapping words closer in proximity and with a higher product of their lengths. Additionally, we conduct manual reviews of the results to ensure coherence and interpretability. Notably, proper nouns are excluded from the swapping process.  The out-of-the-box accuracy for \textbf{\sc{Word-Swap}} on \model is 0.79 (i.e., without fine-tuning on any perturbation set). The model's performance on \textbf{\sc{Word-Swap}} after mix training on all 5 perturbation types, indicating out-of-distribution performance, is summarized in Table~\ref{tab:wordswap}

\input{tables/OOD_WordSwap_table}

\subsubsection{Additional Results on Zero-shot}
\label{sec:appendix_additional_results}

The Table~\ref{table-llm-ins-full} shows zero shot ({\sc OP$_{\text{ZS}}$}) accuracy for different language models.

\begin{table*}[!htb]
\renewcommand{\arraystretch}{1.1}
\centering
\small
\setlength{\tabcolsep}{8pt} % Adjust the spacing between columns
\scalebox{0.98}{
\begin{tabular}{p{0.5cm}|c|ccccc|c@{}}
% \hline
\textbf{Set} & \textbf{Model} & \textbf{char} & \textbf{neg} & \textbf{num} & \textbf{loc} & \textbf{stan} & \textbf{avg.} \\
\hline
\multirow{4}{=}{\begin{sideways}\textbf{ {\sc Unperturbed Q' }}\end{sideways}} 
& Flan-T5-small & 39.30 & 48.60 & 39.30 & 59.60 & 47.00 & 46.76 \\
& Flan-T5-base & 55.60 & 63.60 & 55.60 & 68.00 & 58.60 &  60.28 \\
& Flan-T5-large & 70.60 & 75.00 & 64.60 & 77.00 & 71.60 &  71.76 \\
& Flan-T5-XL & 72.30 & 76.30 & 66.70 & 78.60 & 75.30 & 73.84 \\
& Flan-T5-XXL & 70.60 & 77.30 & 69.00 & 74.00 & 79.00 &  73.98 \\
& LLaMA-2-13b & 51.33 & 54.00 & 49.67 & 62.33 & 53.00 &  54.07 \\
& LLaMA-2-70b & 59.00 & 63.60 & 64.60 & 67.00 & 60.00 &  62.84 \\
& GPT-3.5 & 68.00 & 69.00 & 68.66 & 71.60 & 70.00 &  69.45 \\
\hline
\multirow{4}{=}{\begin{sideways}\textbf{ {\sc Perturbed Q } }\end{sideways}} 
& Flan-T5-small & 33.00 & 40.00 & 49.30 & 71.00 & 47.00 & 48.06 \\
& Flan-T5-base & 44.00 & 54.00 & 55.60 & 68.60 & 58.00 & 56.04 \\
& Flan-T5-large & 54.00 & 66.00 & 62.30 & 65.00 & 67.60 & 62.98 \\
& Flan-T5-XL & 63.00 & 68.00 & 64.00 & 66.00 & 71.30 & 66.46 \\
& Flan-T5-XXL & 63.00 & 70.00 & 63.00 & 65.00 & 69.30 & 66.06 \\
& LLaMA-2-13b & 39.67 & 39.33 & 45.67 & 56.67 & 44.67 & 45.20\\
& LLaMA-2-70b & 54.00 & 51.60 & 49.60 & 57.00 & 54.30 & 53.30 \\
& GPT-3.5 & 51.00 & 53.00 & 62.66 & 61.00 & 60.30 & 57.59 \\
% \hline
\end{tabular}
}
\caption{\textbf{Zero Shot Results ({\sc OP$_{\text{ZS}}$}):} Baseline accuracy for LLMs for Original prompts in zero-shot setting.}

\label{table-llm-ins-full}
\end{table*}

\subsection{Related Works:- Tabular Datasets and Models.} Research on semi-structured tabular data has delved into tasks like tabular natural language inference, fact verification \citep{chen2019tabfact,gupta-etal-2020-infotabs,Zhang:2019:ADC}, and more. Techniques for improving tabular inference include pre-training methods \citep{yu2018spider,yu2020grappa,eisenschlos2020understanding,neeraja-etal-2021-incorporating}. Moreover, recently shared tasks such as SemEval'21 Task 9 \citep{semeval_2021:task9} and FEVEROUS'21 \citep{aly2021feverous} have expanded upon these topics.

\subsection{Implementation Details}\label{sec:appendix:implementationdetails}
\paragraph{For RoBERTA-LARGE}: For creating a baseline model the RoBERTA-LARGE model is fine-tuned on \parentDatasetName for 10 epochs with a learning rate of $1e^{-5}$ with batch size of 4 and adagrad optimizer. \cite{shankarampeta2022enhancing,jain-etal-2021-tabpert}. For fine-tuning on challenge set {\sc P}$_{i}$, we use a learning rate of $3e^{-5}$. This learning is selected after experimenting with various learning rates(specifically $5e^{-4}$, $1e^{-4}$, $5e^{-5}$, $3e^{-5}$, $1e^{-5}$, $5e^{-6}$, $1e^{-6}$) and observing their performance on single set inoculation for various training dataset sizes(specifically $100$, $300$ and $500$). We have used NVIDIA RTX A5000(24 GB), NVIDIA RTX A6000(48 GB) and Google Colab GPU(A100) for conducting different experiments. For the mix fine-tuning we ran the evaluation for 5 different random seeds for each challenge set combination. Average metrics for calculating the final accuracy of mix training to avoid random noise.

\paragraph{{\sc Seq} Metrics.}  \textit{Column Wise Average.} and \textit{Row Wise Average} metrics evaluation:
 
\begin{itemize} 
\item \textit{Column Wise Average.} The column-wise average (COL) for a given perturbation {$\pi_d$} is the average performance improvement over the baseline on {\sc Q}$_j$ (Table~\ref{table-cross-testing-roberta-large}) for models fine-tuned on all other perturbation {\sc P$_j,\text{{ for }} j\neq d$} (except itself). 
   \item \textit{Row Wise Average.} The row-wise average (ROW) for a given perturbation $\pi_d$ is the average performance improvement over the baseline performance (Table~\ref{table-cross-testing-roberta-large}) for the model fine-tuned on {\sc P$_d$} on other challenge dataset sets {\sc Q$_j$, \text{{ for }} $j \neq d$}. 
\end{itemize}

{\sc S$_j$} is sampled randomly from the original dataset {\sc O}. Furthermore, we only consider samples which can be easily perturbed with standard tools such as TextAttack \cite{morris2020textattack}, NLP Checklist \cite{ribeiro-etal-2020-beyond} and manual perturbations supported with paraphrasing tools such as Parrot \cite{zhao-etal-2023-parrot}.  

From {\sc S$_j$} (|{\sc S$_j$}| >= 1500), we sampled {\sc P$_j$} (|{\sc P$_j$}| = 1000) the training perturbation set and {\sc Q$_j$} (|{\sc Q$_j$}| = 500) the testing perturbation set.  To make the sampling diverse and ensure full coverage of the original set, we utilise the Determinantal Point Processes algorithm (DPP) \cite{kulesza2011k}. Determinantal Point Processes (DPPs) are probabilistic models that allow for non-repetitive sampling (diverse \& repulsed) of subsets from a larger set of items. k-DPP is a variant of DPP that conditions the process with a cardinality k, meaning it samples a specific number of items k from the larger set. We use the efficient k-DPP algorithm  \cite{kulesza2011k} for our sampling, k-DPP is a variant of DPP that conditions the process with a cardinality k, meaning it samples a specific number of items k from the larger set.  Note: we ensure that the sample in |{\sc P$_j$}| and |{\sc Q$_j$}| are mutually exclusive.
 
 \textbf{For LLMs}: We used GPT-3.5 model and LLaMA-2 models for our experiments.
GPT-3.5 has been used with a temperature setting of 0.3 (to preserve reproducibility) and 1000 maximum new tokens. LLaMA-2 model has been used after quantization with QLoRA \cite{dettmers2023qlora}, with \textit{nf4} 4-bit quantization.  Double quantization has been employed and \textit{torch.bfloat16} has been used for computations during the quantization. For API calls on GPT-3.5, we have used CPU only. The cost for fine-tuning is: \$0.008 for training,\$0.012 for usage input, \$0.016 for usage output for 1k tokens. The cost for prompting is \$0.008 for 1k tokens. The number of examples are highlighted in the Section~\ref{sec:case_study} and \ref{fine_tune_sec}.

An interesting observation for LLaMA-2 was made which led to the empirical observation that too many examples within the system prompt may also hurt model performance as evidenced from examples \href{https://drive.google.com/file/d/1x4l-3oJMEygCjQQTAk8sFMWoAJLrfS68/view?usp=sharing}{here} and \href{https://drive.google.com/file/d/10V3-ezWwFVtl_ls8qiaugBmYB5gM8ZyF/view?usp=sharing}{here} (\textit{anonymized for submission}). This observation influenced our decision to demonstrate the model using its past conversational history and to limit the system prompt to instructions specific to the model.

For {\sc SEMP}, we utilized three demonstrations from the challenge set and three from the original set. We used six demonstrations for {\sc OP$_{\text{CoT}}$}. We use ten demonstrations for GPT-3.5 in the MESP$_{\text{MPI}}$ setting and fifteen in the MESP$_{\text{MPE}}$ setting. We ensure that for MESP$_{\text{MPI}}$ at least one exemplar is sampled from each perturbation and , for MESP$_{\text{MPE}}$ the brief description captures the core logic of the perturbation.

For LLaMA-2, we used eight demonstrations in MESP$_{\text{MPI}}$ setting and eleven in the MESP$_{\text{MPE}}$ setting. There are minor differences in the NLI Task Explanation for prompts chosen for GPT-3.5 and LLaMA-2 models, these can be found in the corresponding data and examples are given below. This was done as LLaMA-2 performs better with labelling neutral examples as "it is not possible to tell" instead of "neutral". 

For the Flan-T5 series, the model has been pre-trained on the NLI/RTE task. We used the same format for getting the results for zero shot setting ({\sc OP$_{\text{ZS}}$}) as used in \href{https://huggingface.co/google/flan-t5-large?text=Premise%3A++At+my+age+you+will+probably+have+learnt+one+lesson.+Hypothesis%3A++It%27s+not+certain+how+many+lessons+you%27ll+learn+by+your+thirties.+Does+the+premise+entail+the+hypothesis%3F}{Huggingface inference API example} for premise-hypothesis.

For Large Language Model (LLM), we adopted the same selection strategy as for Pre-trained Large Models (PLM, RoBERTa) to select Pj i.e. 500 examples. To select 50 samples, we employed a random uniform sampling method across the set Pj for each perturbation type. Additionally, we chose 50 unperturbed examples totally exclusive (never perturbed) from the original dataset. This resulted in a total training set size of 300 samples. Furthermore, we took meticulous steps to ensure that the samples labelled as 'entailment', 'contradiction', and 'neutral' were evenly balanced across all three categories. 

\vspace{1em}
\noindent \exampleParagraph{Example for {\sc OP$_{\text{ZS}}$} on Flan-T5 series\\}{Premise:  At my age you will probably have learnt one lesson.

~~~~Hypothesis:  It's not certain how many lessons you'll learn by your thirties.

~~~~Does the premise entail the hypothesis?\\}
\vspace{2mm}

\textbf{Fine-Tuning on GPT-3.5: }

The system prompt was provided with the NLI task explaination and mixed perturbation awareness prompt consisting of a brief explanation of all the perturbation types as used in {\sc MESP$_{\text{MPI}}$} for the model gpt-3.5-turbo-0613. The answering scheme does not require an explaination here. A total of 300 samples are used for fine-tuning. Auto hyper-parameters yielded a batch size of 1, 3 epochs and learning rate multiplier of 2\footnote{More details can be found on the \href{https://platform.openai.com/docs/guides/fine-tuning}{openAI documentation} for fine-tuning.}.

An example is given below:
\lstdefinelanguage{json}{
    basicstyle=\small\ttfamily,
    commentstyle=\color{gray},
    showstringspaces=false,
    breaklines=true,
    frame=lines,
    backgroundcolor=\color{white}
    literate=
     *{0}{{{\color{blue}0}}}{1}
      {1}{{{\color{blue}1}}}{1}
      {2}{{{\color{blue}2}}}{1}
      {3}{{{\color{blue}3}}}{1}
      {4}{{{\color{blue}4}}}{1}
      {5}{{{\color{blue}5}}}{1}
      {6}{{{\color{blue}6}}}{1}
      {7}{{{\color{blue}7}}}{1}
      {8}{{{\color{blue}8}}}{1}
      {9}{{{\color{blue}9}}}{1}
      {:}{{{\color{purple}{:}}}}{1}
      {,}{{{\color{purple}{,}}}}{1}
      {\{}{{{\color{brown}{\{}}}}{1}
      {\}}{{{\color{brown}{\}}}}}{1}
      {[}{{{\color{brown}{[}}}}{1}
      {]}{{{\color{brown}{]}}}}{1},
}

\begin{lstlisting}[language=json, caption=Example for fine-tuning GPT-3.5,mathescape]
{
  "messages": [
    {
      "role": "system",
      "content": "In this task, we will ask you to make an inference about the information presented as the premise..."(Prompt containing NLI task description, perturbation awareness and Description of limitation adepted from ${MESP_{MPI}}$  as in GPT-3.5).
    },
    {
       "role": "user",
       "content": "Premise:  The region of WIMA is Worldwide.  WIMA was founded in 1950.  The location of WIMA is the United States.  The website of WIMA is www.wimaworld.com. Hypothesis: WIMA is located in Gambia."."
    },
    {
      "role": "assistant",
      "content": "Answer: No"
    }
  ]
}
\end{lstlisting}

\vspace{-1.5em}
\subsubsection{{\sc\textbf{MESP}} Prompting Example}
Below an \textbf{example prompt for LLaMA-2 for {\sc MESP$_{\text{MPE}}$}}.

\noindent \exampleParagraph{NLI Task Explanation\\}{
In this task, we will ask you to make an inference about the information presented as the premise. We will show you a premise and a hypothesis. Using only the premise and what you believe most people know about the world, you should choose one of the following options for the premise-hypothesis pair: 

1."yes": Based on the information in the premise and what is commonly known, the hypothesis is definitely true, in such a case respond with "yes". 

2."no": Based on the information in the premise and what is commonly known, the hypothesis is definitely false, in such a case respond with "no". 

3."it is not possible to tell": Based on the premise, the hypothesis could be true, but could also be false. We need additional information that is neither commonly known, nor explicitly mentioned in the premise which makes us come to a conclusion. We cannot make an inference about the hypothesis in such a case respond with "it is not possible to tell". 

}

The next part, \textit{perturbation awareness} contains the brief explanation of the respective perturbations. Explanation for one of the perturbation is as below. We have mentioned the prompt for other perturbations later in this section. 

\noindent \exampleParagraph{Perturbation Awareness\\}{
% \vspace{1mm}
\textit{About Typos:}
When labelling sentences based on a premise, it's crucial to recognize and address errors and typos that may occur during hypothesis writing. Typos encompass mistakes like spelling errors and punctuation errors that commonly appear in written content. While numeric typos, involving number replacements, should generally be left uncorrected as they may still make sense in context, character typos, such as misspellings or incorrect word formations, should be corrected to ensure clarity. Maintaining this distinction is essential for preserving hypothesis meaning and readability. It is very important that if you suspect a typo in the hypothesis, attempt correction using premise hints without prompting the user and then attempt to label it yourself.

\textit{About Attention to Numbers:} ... 

\textit{About the Concept of Negation:} ... 

\textit{About Attention to Locations:} ... 

\textit{About Paraphrasing:} ... 

}

\vspace{-1.0em}
\noindent \exampleParagraph{Description of limitation}{ It is critical that you do not use information other than the premise. Take the premise to be ground truth and known to be correct. Use no external knowledge.}

\vspace{-1.0em}
\noindent \exampleParagraph{Answering\\}{
Answer with an explanation in the following format, restricting the answer to only one of the following: "yes" or "no" or "it is not possible to tell"

E: <explanation>  

A: <answer>

}

\vspace{1mm}
There are multiple \textit{demonstrations} based on the method. We have specified the number of demonstrations used in the implementation details section. In case of the MESP, the demonstrations contains instance of unperturbed as well as perturbed hypothesis NLI tasks. A single instance of a demonstration is shown below, see\textbf{Demostrations}:
\vspace{1mm}

We have shown the prompt in the raw text format but depending on the model the prompt may be changed to adapt to the model's specific behaviour. For example in case of LLaMA-2 model, the NLI task explanation, Perturbation awareness and Description of limitation section are provided as the system prompt, which is consistent with the paper \citealt{touvron2023llama}.

The only difference between {\sc MESP$_{\text{MPE}}$} and {\sc MESP$_{\text{MPI}}$} is that the former has more number of CoT examples of each perturbation in the demonstration section whereas the later has more detailed description of each perturbation in the perturbation awareness section. The perturbation awareness for each type of perturbation for both of the method is at the end of this section.

\noindent \exampleParagraph{Demonstrations\\}{
Premise:  The official languages of Hong Kong Special Administrative Region of the People's Republic of China are Chinese, English.  The regional language of Hong Kong Special Administrative Region of the People's Republic of China is Cantonese.  The official scripts of Hong Kong Special Administrative Region of the People's Republic of China are Traditional Chinese, English alphabet.  The government of Hong Kong Special Administrative Region of the People's Republic of China is Devolved executive-led system within a socialist republic. 
\\Hypothesis: The Hong Kong Special Administrative Region of the People's Republic of China grants official status to more than one language. 
\\ E: To make an inference about the hypothesis, we need to either know directly or deduce how many languages are official in Hong Kong Special Administrative Region of the People's Republic of China. We can see in the premise that There are two official languages: English and Chinese. As the hypothesis says "more than one". As two is more than one, the answer is Yes.
\\A: yes 

Premise: ...\\
Hypothesis: ...\\
E: ...\\
A: ...\\
.\\
.\\
.

}

\subsubsection{{\sc \textbf{SEMP}} Prompting}
% \vspace{-0.75em}
For the {\sc SEMP} method, the perturbation awareness section contains only description of only one kind of perturbation adapted from the \textit{perturbation awareness} section as in {\sc MESP$_{\text{MPI}}$} and the demonstration section contains demonstrations of only one type of perturbation demonstration and with unperturbed demonstrations.

% \vspace{-0.75em}
\subsubsection{{\sc \textbf{OP$_{\text{ZS}}$}} Prompting}
% \vspace{-0.75em}

In case of zero-shot prompting we only explain the NLI task to the model briefly and provide it with the answering format. We have provided example of \textbf{{\sc OP$_{\text{ZS}}$}} below as used in \textbf{GPT-3.5}:

\vspace{1.0em}
\noindent \exampleParagraph{NLI Task Explanation for GPT-3.5\\}
{

In this task, we will ask you to make an inference about the information presented as the premise. We will show you a premise and a hypothesis.
Using only the premise and what you believe most people know about the world, you should choose one of the following options for the premise-hypothesis pair:

Based on the information in the premise and what is commonly known, the hypothesis is definitely true, in such a case respond with Yes.

Based on the information in the premise and what is commonly known, the hypothesis is definitely false, in such a case respond with No.

Based on the premise, the hypothesis could be true, but could also be false. We need additional information that is neither commonly known, nor explicitly mentioned in the premise which makes us come to a conclusion, in such a case respond with Neutral.

}

In the \textbf{{\sc OP$_{\text{ZS}}$}} the perturbation awareness part is not given. So, model is not made aware of any perturbations explicitly.

\noindent \exampleParagraph{Description of limitation\\}{
Avoid using information that you may know if you believe that it is not generally known.

}
\noindent \exampleParagraph{Answering\\}{
Now classify the following Premise-Hypothesis pair. Answer only with one word: Yes or No or Neutral.

}

As this is the zero-shot prompting no demonstration is provided.

\subsubsection{{{\sc \textbf{OP$_{\text{CoT}}$}}} Prompting}
In case of the few-shot with CoT prompting(\textbf{{\sc OP$_{\text{CoT}}$}}), we will also provide examples of the NLI task on unperturbed examples along with its chain of thought explanation as a part of demonstrations. The prompt for \textbf{{\sc OP$_{\text{CoT}}$}} on \textbf{GPT-3.5}. 

\noindent \exampleParagraph{NLI Task Explanation\\}{
\textit{Same as in for {\sc OP$_{\text{ZS}}$}.}
}

Note, that there is no perturbation awareness for CoT prompts.

\noindent \exampleParagraph{Description of limitation\\}{
It is very important and critical that you do not use information other than the premise that you may know if you believe that it is not generally known. 
This restriction should not prevent you from exploring the premise repeatedly and making some assumptions and deeper inferences from the information within the premise.

}
\noindent \exampleParagraph{Demonstration\\}{
Here are some examples:

Premise:    Jerusalem is a city.  The jewish of Jerusalem is 64\%.  The time zone of Jerusalem is UTC+02:00 (IST, PST).  The area code of Jerusalem is +972-2. 

Hypothesis: Christians comprise a big part of the population of Jerusalem.

To make an inference about the hypothesis, we need to either know directly or deduce the population division in Jerusalem. As stated in the premise, Jewish (religion) constitutes 64 percent of the population in Jerusalem. Hence the hypothesis must be false as the Christians(religion) can't possibly constitute a big part of the population, as the majority is taken up by the Jewish. The answer is No.

Premise: ...

Hypothesis: ...

CoT with answer: ...
.
}

Note that in all of the methods the premise-hypothesis pair for NLI task will be at the end of the prompt which will be appended with the shown prompt of each method.

\vspace{-1.5em}
\subsubsection{Detailed perturbation awareness prompts}
\vspace{-0.5em}

\noindent \textbf{Prompts for perturbation awareness {\sc MESP$_{\text{MPI}}$}}:\\
% Find below the prompt for \textit{perturbation awareness} description for different perturbations:
\noindent \exampleParagraph{Perturbation Awareness}{
\textit{\\About typos:}
When performing a labelling task on sentences based on a premise, it's important to understand that errors and typos can occur during the writing of questions. Typos are mistakes made when typing or printing, which can include spelling errors and punctuation errors. These errors can commonly appear in written content and can sometimes affect the clarity and accuracy of a question. The concept of numeric and character typos in questions is important for maintaining the integrity and meaning of a sentence or premise: Numeric typos, where a number is accidentally replaced by another number, should generally not be corrected. This is because the new number may still make sense in the context and altering it could change the question's meaning significantly. It's crucial to recognize that the typo might convey a different question altogether. On the other hand, character typos, such as misspellings or incorrect word formations, should be corrected. These typos often result in words that have no meaning or make the question unclear. Correcting character-based typos is essential to ensure the question remains coherent and can be understood by the reader. Maintaining this distinction is vital for ensuring that the question retains its intended meaning and readability. Numeric typos, although errors, can sometimes add unique value to a question, whereas character typos usually hinder comprehension and should be rectified whenever possible. While numeric typos (errors in numbers) may not always need correction, character-based typos (errors in letters or characters) should be corrected. Numeric typos when a number is replaced by another number, shouldn't be corrected as this can mean a different question altogether where the new number still makes sense. Character typos where the newly formed word (after a typo) has no meaning, should be corrected and attempted to be reformed to the original word hints of the original word may also be made from the premise. The reason typos happen during typing is because our brains focus on conveying meaning rather than the fine details of individual characters. This phenomenon can lead to errors slipping through. In a labelling task, it's crucial to be vigilant about character-based typos as they can affect the interpretation of the premise and the accuracy of labelling. 

}

\noindent \exampleParagraph{}{
\textit{About attention to locations:}
Here is some additional information which may help. Prioritize Location Accuracy: In this labelling task, it is of utmost importance to ensure the precise handling of location-related information. Pay close attention to locations and prioritize accuracy over other details. Use Abbreviations and Basic General Knowledge: Allow for the use of abbreviations like "NY" (New York) or "IND" (Indianapolis or India either may work depending on context). Basic general knowledge about locations, such as their geographical features and neighboring regions, is acceptable. However, do not include historical facts or general events about the place. Verify with External Resources: Encourage the utilization of external resources for verification when dealing with critical location data. Whenever possible, cross-reference the provided information with reliable sources such as maps, atlases, or official websites to ensure correctness. Review and Edit Meticulously: Emphasize the importance of reviewing and editing location-related responses meticulously before finalizing the answer. Double-check the spelling, coordinates, and other location-specific details to guarantee precision.

}

\noindent \exampleParagraph{}{
\textit{About attention to numbers:}
Please pay meticulous attention to numerical information. When performing labelling tasks, it is crucial to handle numerical data with precision. Ensure that the responses contain specific numerical values and context. Emphasize the importance of self-rechecking critical numerical information, and remind yourself to thoroughly review and edit numerical responses for accuracy before finalizing the answer.

In labelling tasks, the hypotheses may contain numerical values. When encountering such cases, carefully identify the numerical data and ensure that it is accurately labelled. Pay close attention to the context and surrounding words as well as arithmetic operators (e.g., +, -, *, /) that may influence the meaning of the numerical value.

Your goal is to provide labels that infer the answer from correct numerical values and comparisons and also reflect the nuanced inferences made from the presence of more or less types of words and arithmetic operators. This entails understanding the role of numerical data in the context of the hypothesis and accurately capturing its significance in the labels.

Remember that precision and accuracy in handling numerical information are paramount in labelling tasks. Take your time to review and edit your numerical responses, double-checking for any potential errors or omissions to ensure the highest quality labelling results.

}

\noindent \exampleParagraph{}{
\textit{About paraphrasing:}
When performing a labelling task where you need to analyze a sentence or a piece of text, it's crucial to understand that the question posed may not always be presented in the exact same words as the information you are reading. This is where the concept of paraphrasing comes into play.

Paraphrasing involves rephrasing a sentence or passage while retaining its original meaning. It's a common practice in various contexts, including academic writing, as it allows for the expression of the same idea in different words. Paraphrasing can help you better understand and articulate information, and it's especially important when dealing with labelling tasks where the wording might not match exactly.

In the context of a labelling task, you should be aware that the question you're trying to answer might be a paraphrased version of the information presented in the text or a sentence in the premise. This paraphrasing may not be perfect, and there could be slight variations or synonyms used. Therefore, it's essential to carefully read and analyze the text, looking for similarities in meaning rather than relying solely on identical phrasing. By doing so, you can effectively identify and label the relevant information, even if it's not presented verbatim. Paraphrasing skills are valuable in such tasks as they allow you to recognize the core concepts and convey them accurately, regardless of the wording used in the question.

If you feel like the hypothesis may have a typo, you should attempt to correct it yourself by taking hints from the premise to guess the actual hypothesis and then attempt to label it. Do not prompt the user to correct the hypothesis, attempt it yourself.

}

\noindent \exampleParagraph{}{
\textit{About the concept of negation:}
It may also be necessary to understand the concept of negation to make correct inferences. Negation in sentences is the process of expressing the opposite or denial of something. When someone has to pay close attention to statements, understanding negation is crucial because it can change the meaning of a sentence significantly.

Single Negation: In a sentence with a single negation, a negative word like "not" or "no" is used to express a negative statement. For example, "I do not like ice cream" means the person dislikes ice cream.

Double Negation: While less commonly used than single negation, this occurs when two negative words are used in a sentence, such as "I don't want no ice cream." In this case, the double negative creates an affirmative or positive meaning, so the sentence means "I want ice cream.

Triple Negation: While used very rarely, triple negation involves the use of three negative words in a sentence, like "I don't need no help." In this case, it also conveys a positive meaning, indicating that the person doesn't require any assistance.

For someone paying close attention to statements, it's essential to recognize double or triple negations to accurately understand the speaker's intended meaning. These constructions often appear in colloquial speech, so close attention to context and word usage is necessary to avoid misinterpretation.

}

\textbf{All prompts for perturbation awareness for {\sc MESP$_{\text{MPE}}$}}:\\
Find below the prompt for \textit{perturbation awareness} description for different perturbations.

\noindent \exampleParagraph{Perturbation Awareness\\}{
\textit{About Typos:}
\textit{already shown in the MESP prompt.}
}
\noindent \exampleParagraph{}{
\textit{About Attention to Numbers:}
Precise handling of numerical information is paramount in labelling tasks. Be diligent in ensuring numerical data accuracy, considering context, surrounding words, and arithmetic operators. Labels should reflect nuanced inferences drawn from numerical values and word usage.
It is very important to recheck numeric calculations and arithmetic and mathematical operations. 

}
\noindent \exampleParagraph{}{
\textit{About the Concept of Negation:}
Understanding negation is crucial as it can significantly alter sentence meaning. Single negation involves using negative words like "not" to express negativity, while double negation can turn a negative statement into a positive one. Triple negation is rare but also conveys a positive meaning. Close attention to context is essential to avoid misinterpretation. 

}
\noindent \exampleParagraph{}{
\textit{About Attention to Locations:}
Location accuracy is a top priority in labelling tasks. Use abbreviations and basic location knowledge, but avoid historical facts. Verify location data with external resources when critical. Meticulously review and edit location-related responses for precision.

}
\noindent \exampleParagraph{}{
\textit{About Paraphrasing:}
In labelling tasks, hypotheses may not mirror the premise's wording exactly. Paraphrasing, or rephrasing with the same meaning, is common. Carefully analyze premise for similar meanings and core concepts, even if phrasing varies. Paraphrasing skills help identify and label relevant information accurately.  

}

\subsection{LLM Answer Extraction Module}
\begin{figure}[!h]
    \centering
\includegraphics[width=0.95\linewidth]{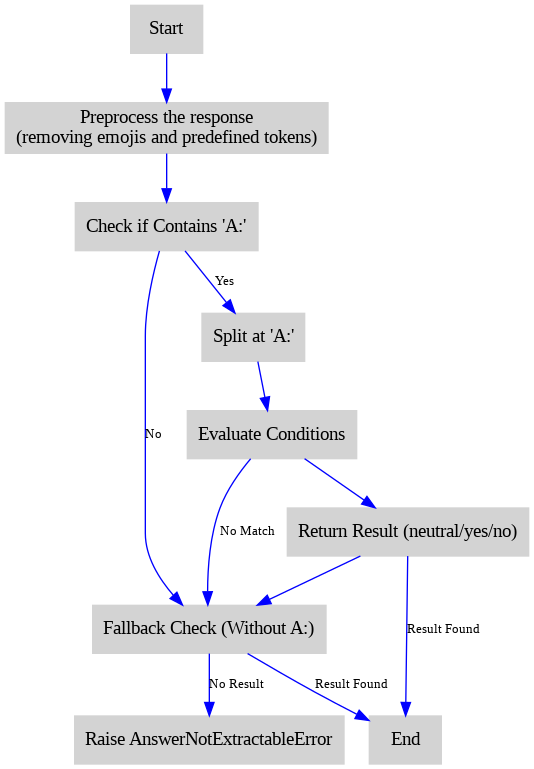}
\vspace{-0.5em}
    \caption{\textbf{Flowchart for answer extraction}  }
    \label{fig:answer_extractor}
    \vspace{-1.0em}
\end{figure}

% \vk{Finish this up creata layman description for extraction as an example from the given answer.}
The outputs of the large language models are not necessarily in the required format even after explicitly specifying the format. Thus, we needed to design a method to extract out the answer from the very verbose outputs of the model. So, we have shown the flow of the answer extraction module in the Fig~\ref{fig:answer_extractor}.
The module begins by removing non-essential elements such as emojis from the text, enhancing text clarity for analysis. It then searches for a key marker (`A:'), indicating the start of a relevant response. Upon identification, this section is isolated for evaluation.

The module's functionality is centered on categorizing responses into affirmative, negative, or neutral based on specific phrases. In cases where the marker is missing, it reassesses the entire text, ensuring comprehensive analysis. If the response remains ambiguous, the module raises an error.

\subsection{Confusion Graphs} 
The confusion graph below represents the confusion matrix values for char, neg, num, loc, stan perturbation for a particular method in the results section. This results provide the insights on which type of hypothesis out of entailment, contradiction and neutral are more difficult for the model with given method. The arrow from A to B represents the percentage of examples which has true label A and has been predicted as B. All the graphs are on perturbed sets.

\begin{figure}
\centering
\begin{minipage}{.45\textwidth}
  \centering
  {\small
  \input{consistency\_MESP\_MPE\_GPT.tikz}}
    \captionof{figure}{Confusion graph {\sc MESP$_{\text{MPE}}$} for  GPT-3.5 on char, neg, num, loc and stan respectively.}
    \label{fig:consistency_graph:1}
\end{minipage}%

\begin{minipage}{.45\textwidth}
  \centering
{\small \input{consistency\_MESP\_MPI\_GPT.tikz}}
    \captionof{figure}{ Confusion graph {\sc MESP$_{\text{MPI}}$} for   GPT-3.5 on char, neg, num, loc and stan respectively.}
    \label{fig:consistency_graph:2}
\end{minipage}

\begin{minipage}{.48\textwidth}
  \centering
   {\small \input{consistency\_SEMP\_char\_aware\_GPT.tikz}}
    \captionof{figure}{Confusion graph {\sc SEMP$_{\text{char}}$} for   GPT-3.5 on char, neg, num, loc and stan respectively.}
    \label{fig:consistency_graph:3}
    \smallskip
\end{minipage}%
\end{figure}

\begin{figure}
\centering
\begin{minipage}{.48\textwidth}
  \centering
  {\small \input{consistency\_SEMP\_neg\_aware\_GPT.tikz}}
    \captionof{figure}{Confusion graph {\sc SEMP$_{\text{neg}}$} for   GPT-3.5 on char, neg, num, loc and stan respectively.}
    \label{fig:consistency_graph:4}
    \smallskip
\end{minipage}

\begin{minipage}{.48\textwidth}
  \centering
{\small \input{consistency\_SEMP\_num\_aware\_GPT.tikz}}
    \captionof{figure}{Confusion graph {\sc SEMP$_{\text{num}}$} for   GPT-3.5 on char, neg, num, loc and stan respectively.}
    \label{fig:consistency_graph:5}
    \smallskip
\end{minipage}%

\begin{minipage}{.48\textwidth}
  \centering
  {\small \input{consistency\_SEMP\_loc\_aware\_GPT.tikz}}
    \captionof{figure}{Confusion graph {\sc SEMP$_{\text{loc}}$} for   GPT-3.5 on char, neg, num, loc and stan respectively.}
    \label{fig:consistency_graph:6}
    \smallskip
\end{minipage}
\end{figure}

\begin{figure}
\centering
\begin{minipage}{.48\textwidth}
  \centering
  {\small \input{consistency\_SEMP\_stan\_aware\_GPT.tikz}}
    \captionof{figure}{Confusion graph {\sc SEMP$_{\text{stan}}$} for   GPT-3.5 on char, neg, num, loc and stan respectively.}
    \label{fig:consistency_graph:7}
    \smallskip
\end{minipage}%

\begin{minipage}{.48\textwidth}
  \centering
  {\small \input{consistency\_OP\_ZS\_GPT.tikz}}
    \captionof{figure}{Confusion graph {\sc OP$_{\text{ZS}}$} for   GPT-3.5 on char, neg, num, loc and stan respectively.}
    \label{fig:consistency_graph:8}
    \smallskip
\end{minipage}

\begin{minipage}{.48\textwidth}
  \centering
  {\small \input{consistency\_OP\_COT\_GPT.tikz}}
    \captionof{figure}{Confusion graph {\sc OP$_{\text{CoT}}$} for   GPT-3.5 on char, neg, num, loc and stan respectively.}
    \label{fig:consistency_graph:9}
    \smallskip
\end{minipage}%
\end{figure}

\begin{figure}
\centering
\begin{minipage}{.48\textwidth}
  \centering
  {\small \input{consistency\_SEQ\_column\_DSC.tikz}}
    \captionof{figure}{Confusion graph {\sc SEQ$_{\text{COL-ASC}}$} for   \model on char, neg, num, loc and stan respectively.}
    \label{fig:consistency_graph:10}
    \smallskip
\end{minipage}

\centering
\begin{minipage}{.48\textwidth}
  \centering
  {\small \input{consistency\_MIX\_500.tikz}}
    \captionof{figure}{Confusion graph {\sc MIX} with 500 examples each for  \model on char, neg, num, loc and stan respectively.}
    \label{fig:consistency_graph:11}
    \smallskip
\end{minipage}%

\begin{minipage}{.48\textwidth}
  \centering
{\small \input{consistency\_DYNMIX\_1500.tikz}}
    \captionof{figure}{Confusion graph {\sc DYNMIX} with total 1500 examples  for  \model on char, neg, num, loc and stan respectively.}
    \label{fig:consistency_graph:12}
    \smallskip
\end{minipage}
\end{figure}

%% file: tables/in-out-merge.tex
\begin{table*}[!ht]
    \centering
    \small
    \setlength{\tabcolsep}{6pt} % Adjust the spacing between columns
    \begin{subtable}{\linewidth}
        \centering
        \begin{tabular}{l|ccc|cc|ccc|c}
           % \cline{1-10}
            & \multicolumn{3}{c|}{\textbf{In-distribution}} & \multicolumn{2}{c|}{\textbf{Out-distribution}} & \multicolumn{3}{c|}{\textbf{Original Test sets}} & \\
            \cline{1-10}
            \textbf{K} & \textbf{neg} & \textbf{num} & \textbf{loc} & \textbf{char} & \textbf{stan} & \textbf{alpha1} & \textbf{alpha2} & \textbf{alpha3} & \textbf{$\mu$} \\
            \cline{1-10}
            \textbf{baseline} & 46.90 & 67.20 & 70.20 & \bf 57.30 & \bf 67.10 & \bf 72.72 & \bf 64.83 & \bf 62.33 & - \\
            \hline
            \textbf{100} & 60.4 & 83.2 & 81.4 & 49.6 & 59.6 & 63.6 & 62.8 & 56.1 & 5.10 \\
            \textbf{200} & 61.9 & 85.6 & 83.0 & 49.2 & 58.0 & 61.3 & 61.9 & 53.0 & 5.79 \\
            \textbf{300} & 62.1 & 85.8 & 83.2 & 48.8 & 55.7 & 59.4 & 62.3 & 51.9 & 5.39 \\
            \textbf{400} & 66.3 & 85.1 & 83.5 & 47.5 & 54.3 & 58.4 & 61.5 & 51.1 & 5.61 \\
            \textbf{500} & \bf 68.0 & \bf 86.0 & \bf 84.1 & 47.8 & 53.9 & 58.0 & 61.2 & 50.1 & \bf 6.23 \\
            %\cline{1-10}
        \end{tabular}
        \caption{\textbf{Fine Tuning on Perturbation Subset (neg, num, loc).} Model fine tuned using {\sc Mix} strategy using only 3 perturbations. Performance reported on out of distribution perturbation and alpha test sets.}
        \label{tab:in-out-1}
    \end{subtable}

    \vspace{0.3cm} % Add some vertical space between the tables

    \begin{subtable}{\linewidth}
        \centering
        \begin{tabular}{l|cc|ccc|ccc|c}
            %\cline{1-10}
            & \multicolumn{2}{c|}{\textbf{In-distribution}} & \multicolumn{3}{c|}{\textbf{Out-distribution}} & \multicolumn{3}{c|}{\textbf{Original Test sets}} & \\
            \cline{1-10}
            \textbf{K} & \textbf{char} & \textbf{num} & \textbf{neg} & \textbf{loc} & \textbf{stan} & \textbf{alpha1} & \textbf{alpha2} & \textbf{alpha3} & \textbf{$\mu$} \\
            \cline{1-10}
            \textbf{baseline} & 57.30 & 67.20 & 46.90 & 70.20 & 67.10 & \bf 72.72 & \bf 64.83 & \bf 62.33 & - \\
            \hline
            \textbf{100} & 56.3 & 80.1 & 50.3 & 74.6 & 65.4 & 71.0 & 63.2 & 60.1 & 3.61 \\
            \textbf{200} & 57.2 & 82.8 & 47.9 & 76.3 & 65.3 & 70.9 & 63.5 & 59.2 & 4.15 \\
            \textbf{300} & 57.0 & 83.1 & 47.0 & 77.1 & 65.2 & 71.1 & 63.1 & 58.1 & 4.13 \\
            \textbf{400} & \bf 58.0 & \bf 84.1 & \bf 48.5 & \bf 78.0 & \bf 64.4 & 70.8 & 63.8 & 58.4 & \bf 4.86 \\
            \textbf{500} & 57.0 & \bf 84.1 & 46.7 & 77.7 & \bf 64.4 & 70.9 & 63.2 & 58.0 & 4.25 \\
            %\cline{1-10}
        \end{tabular}
        \caption{\textbf{Fine Tuning on Perturbation Subset (char, num).} Model fine tuned using {\sc Mix} strategy using only 2 perturbations. Performance reported on out of distribution perturbation and alpha test sets.}
        \label{tab:in-out-2}
    \end{subtable}
    \caption{In-distribution represents perturbation types used for training, Out-distribution are the other perturbation types. \textbf{K} is the number of samples used for each perturbation during training. \textbf{$\mu$} is the average improvement over the baseline of all perturbation sets.}
    \label{tab:common-in-out}
\end{table*}

%% file: tables/OOD_WordSwap_table.tex
\begin{table}[h!]
\centering
\small
\begin{tabular}{ c|c|c|c|c|c }
% \hline
 K & 100 & 200 & 300 & 400 & 500 \\ 
\hline
 Acc. & 73.4 & 73.2 & 71.6 & 74.0 & 74.6 \\ 
% \hline
\end{tabular}
\caption{Performance of model on \textsc{Word-Swap} Perturbation with MIX training. Acc. is the accuracy on \textsc{Word-Swap} type perturbation and K is the number of samples.}\label{tab:wordswap}
\end{table}

%% file: consistency_MESP_MPE_GPT.tikz
    \begin{tikzpicture}[thick, scale=0.65, {every node/.style}={scale=0.65}]
        \begin{pgfonlayer}{nodelayer}
            % Nodes
            \node [style=new style 0] (0) at (0, 8) {C};
            \node [style=new style 0] (1) at (-7, 0) {E};
            \node [style=new style 0] (2) at (7, 0) {N};
            \node [style=none] (3) at (-6.5, 1.5) {};
            \node [style=none] (4) at (-6, 1) {};
            \node [style=none] (5) at (-5.75, -0.5) {};
            \node [style=none] (6) at (-5.75, 0.25) {};
            \node [style=none] (7) at (-1.25, 7.25) {};
            \node [style=none] (8) at (-0.75, 6.75) {};
            \node [style=none] (9) at (0.75, 6.75) {};
            \node [style=none] (10) at (1.25, 7.25) {};
            \node [style=none] (11) at (6, 1) {};
            \node [style=none] (12) at (6.5, 1.5) {};
            \node [style=none] (13) at (5.5, -0.5) {};
            \node [style=none] (14) at (5.5, 0.25) {};
            \node [style=new style 1] (20) at (0, 11.75) {};

            % Values
            \node [style=new style 1] at (7.5, -3.75) {20.3, 24.7, 11.0, 32.7, 23.3}; % N to N
            \node [style=new style 1] at (-7.25, -3.75) {31.0, 10.7, 3.7, 0.0, 23.0}; % E to E
            \node [style=new style 1] at (0, 11.75) {18.7, 28.3, 43.7, 27.3, 20.3}; % C to C
            \node [style=new style 2, rotate=47] at (-4.75, 4.5) {2.3, 13.3, 0.0, 0.0, 3.7}; % E to C
            \node [style=new style 2, rotate=47] at (-3, 3.25) {7.3, 2.0, 13.0, 2.0, 4.0}; % C to E
            \node [style=new style 2] at (0, -1.25) {7.3, 1.0, 1.3, 0.7, 9.0}; % N to E
            \node [style=new style 2] at (0, 1) {3.0, 9.3, 1.3, 0.0, 5.3}; % E to N
            \node [style=new style 2, rotate=-47] at (4.25, 5) {1.7, 7.0, 10.0, 11.7, 1.7}; % N to C
            \node [style=new style 2, rotate=-47] at (2.75, 3.5) {8.3, 3.7, 16.0, 25.7, 9.7}; % C to N
        \end{pgfonlayer}
        \begin{pgfonlayer}{edgelayer}
            		\draw [style=new edge style 0] (8.center) to (4.center); % Entail and Contradiction Top arrow
                \draw [style=new edge style 0] (14.center) to (6.center); % Entail and Neutral Top arrow
                \draw [style=new edge style 3, in=-135, out=-45, loop] (2) to ();
                \draw [style=new edge style 0] (5.center) to (13.center); %Entail and Neutral bottom arrow 
                \draw [style=new edge style 0] (3.center) to (7.center); %Entail and contra bottom arrow
                \draw [style=new edge style 0] (9.center) to (11.center); 
                \draw [style=new edge style 0] (12.center) to (10.center);
                \draw [style=new edge style 3, in=-135, out=-45, loop] (1) to ();
                \draw [style=new edge style 3, in=135, out=45, loop] (0) to ();
        \end{pgfonlayer}
    \end{tikzpicture}
    

%% file: consistency_MESP_MPI_GPT.tikz
    \begin{tikzpicture}[thick, scale=0.65, {every node/.style}={scale=0.65}]
        \begin{pgfonlayer}{nodelayer}
            % Nodes
            \node [style=new style 0] (0) at (0, 8) {C};
            \node [style=new style 0] (1) at (-7, 0) {E};
            \node [style=new style 0] (2) at (7, 0) {N};
            \node [style=none] (3) at (-6.5, 1.5) {};
            \node [style=none] (4) at (-6, 1) {};
            \node [style=none] (5) at (-5.75, -0.5) {};
            \node [style=none] (6) at (-5.75, 0.25) {};
            \node [style=none] (7) at (-1.25, 7.25) {};
            \node [style=none] (8) at (-0.75, 6.75) {};
            \node [style=none] (9) at (0.75, 6.75) {};
            \node [style=none] (10) at (1.25, 7.25) {};
            \node [style=none] (11) at (6, 1) {};
            \node [style=none] (12) at (6.5, 1.5) {};
            \node [style=none] (13) at (5.5, -0.5) {};
            \node [style=none] (14) at (5.5, 0.25) {};
            \node [style=new style 1] (20) at (0, 11.75) {};

            % Values
            \node [style=new style 1] at (7.5, -3.75) {15.7, 24.3, 10.3, 26.7, 16.0}; % N to N
            \node [style=new style 1] at (-7.25, -3.75) {19.3, 12.7, 3.3, 0.0, 23.3}; % E to E
            \node [style=new style 1] at (0, 11.75) {24.7, 28.3, 48.3, 32.3, 20.3}; % C to C
            \node [style=new style 2, rotate=47] at (-4.75, 4.5) {11.3, 11.3, 0.7, 0.0, 4.7}; % E to C
            \node [style=new style 2, rotate=47] at (-3, 3.25) {3.3, 1.3, 11.3, 3.0, 4.3}; % C to E
            \node [style=new style 2] at (0, -1.25) {6.3, 0.3, 1.7, 1.0, 14.0}; % N to E
            \node [style=new style 2] at (0, 1) {5.7, 9.3, 1.0, 0.0, 4.0}; % E to N
            \node [style=new style 2, rotate=-47] at (4.25, 5) {7.3, 8.0, 10.3, 17.3, 4.0}; % N to C
            \node [style=new style 2, rotate=-47] at (2.75, 3.5) {6.3, 4.3, 13.0, 19.7, 9.3}; % C to N
        \end{pgfonlayer}
        \begin{pgfonlayer}{edgelayer}
            		\draw [style=new edge style 0] (8.center) to (4.center); % Entail and Contradiction Top arrow
                \draw [style=new edge style 0] (14.center) to (6.center); % Entail and Neutral Top arrow
                \draw [style=new edge style 3, in=-135, out=-45, loop] (2) to ();
                \draw [style=new edge style 0] (5.center) to (13.center); %Entail and Neutral bottom arrow 
                \draw [style=new edge style 0] (3.center) to (7.center); %Entail and contra bottom arrow
                \draw [style=new edge style 0] (9.center) to (11.center); 
                \draw [style=new edge style 0] (12.center) to (10.center);
                \draw [style=new edge style 3, in=-135, out=-45, loop] (1) to ();
                \draw [style=new edge style 3, in=135, out=45, loop] (0) to ();
        \end{pgfonlayer}
    \end{tikzpicture}
    

%% file: consistency_SEMP_char_aware_GPT.tikz
    \begin{tikzpicture}[thick, scale=0.65, {every node/.style}={scale=0.65}]
        \begin{pgfonlayer}{nodelayer}
            % Nodes
            \node [style=new style 0] (0) at (0, 8) {C};
            \node [style=new style 0] (1) at (-7, 0) {E};
            \node [style=new style 0] (2) at (7, 0) {N};
            \node [style=none] (3) at (-6.5, 1.5) {};
            \node [style=none] (4) at (-6, 1) {};
            \node [style=none] (5) at (-5.75, -0.5) {};
            \node [style=none] (6) at (-5.75, 0.25) {};
            \node [style=none] (7) at (-1.25, 7.25) {};
            \node [style=none] (8) at (-0.75, 6.75) {};
            \node [style=none] (9) at (0.75, 6.75) {};
            \node [style=none] (10) at (1.25, 7.25) {};
            \node [style=none] (11) at (6, 1) {};
            \node [style=none] (12) at (6.5, 1.5) {};
            \node [style=none] (13) at (5.5, -0.5) {};
            \node [style=none] (14) at (5.5, 0.25) {};
            \node [style=new style 1] (20) at (0, 11.75) {};

            % Values
            \node [style=new style 1] at (7.5, -3.75) {20.3, 23.0, 8.3, 22.3, 17.7}; % N to N
            \node [style=new style 1] at (-7.25, -3.75) {29.7, 12.7, 3.7, 0.0, 24.0}; % E to E
            \node [style=new style 1] at (0, 11.75) {17.7, 29.7, 54.0, 39.0, 21.7}; % C to C
            \node [style=new style 2, rotate=47] at (-4.75, 4.5) {3.3, 12.7, 0.0, 0.0, 5.0}; % E to C
            \node [style=new style 2, rotate=47] at (-3, 3.25) {7.3, 2.0, 8.7, 2.7, 5.7}; % C to E
            \node [style=new style 2] at (0, -1.25) {7.3, 0.3, 3.0, 1.0, 10.7}; % N to E
            \node [style=new style 2] at (0, 1) {3.3, 8.0, 1.3, 0.0, 3.0}; % E to N
            \node [style=new style 2, rotate=-47] at (4.25, 5) {1.7, 9.3, 11.0, 21.7, 5.7}; % N to C
            \node [style=new style 2, rotate=-47] at (2.75, 3.5) {9.3, 2.3, 10.0, 13.3, 6.7}; % C to N
        \end{pgfonlayer}
        \begin{pgfonlayer}{edgelayer}
            		\draw [style=new edge style 0] (8.center) to (4.center); % Entail and Contradiction Top arrow
                \draw [style=new edge style 0] (14.center) to (6.center); % Entail and Neutral Top arrow
                \draw [style=new edge style 3, in=-135, out=-45, loop] (2) to ();
                \draw [style=new edge style 0] (5.center) to (13.center); %Entail and Neutral bottom arrow 
                \draw [style=new edge style 0] (3.center) to (7.center); %Entail and contra bottom arrow
                \draw [style=new edge style 0] (9.center) to (11.center); 
                \draw [style=new edge style 0] (12.center) to (10.center);
                \draw [style=new edge style 3, in=-135, out=-45, loop] (1) to ();
                \draw [style=new edge style 3, in=135, out=45, loop] (0) to ();
        \end{pgfonlayer}
    \end{tikzpicture}
    

%% file: consistency_SEMP_neg_aware_GPT.tikz
    \begin{tikzpicture}[thick, scale=0.65, {every node/.style}={scale=0.65}]
        \begin{pgfonlayer}{nodelayer}
            % Nodes
            \node [style=new style 0] (0) at (0, 8) {C};
            \node [style=new style 0] (1) at (-7, 0) {E};
            \node [style=new style 0] (2) at (7, 0) {N};
            \node [style=none] (3) at (-6.5, 1.5) {};
            \node [style=none] (4) at (-6, 1) {};
            \node [style=none] (5) at (-5.75, -0.5) {};
            \node [style=none] (6) at (-5.75, 0.25) {};
            \node [style=none] (7) at (-1.25, 7.25) {};
            \node [style=none] (8) at (-0.75, 6.75) {};
            \node [style=none] (9) at (0.75, 6.75) {};
            \node [style=none] (10) at (1.25, 7.25) {};
            \node [style=none] (11) at (6, 1) {};
            \node [style=none] (12) at (6.5, 1.5) {};
            \node [style=none] (13) at (5.5, -0.5) {};
            \node [style=none] (14) at (5.5, 0.25) {};
            \node [style=new style 1] (20) at (0, 11.75) {};

            % Values
            \node [style=new style 1] at (7.5, -3.75) {20.7, 24.7, 11.7, 31.3, 23.0}; % N to N
            \node [style=new style 1] at (-7.25, -3.75) {22.0, 12.3, 3.3, 0.0, 22.7}; % E to E
            \node [style=new style 1] at (0, 11.75) {17.7, 27.7, 43.3, 28.3, 17.7}; % C to C
            \node [style=new style 2, rotate=47] at (-4.75, 4.5) {7.7, 11.3, 0.3, 0.0, 4.7}; % E to C
            \node [style=new style 2, rotate=47] at (-3, 3.25) {3.0, 3.0, 9.7, 2.3, 4.0}; % C to E
            \node [style=new style 2] at (0, -1.25) {6.0, 0.3, 1.7, 1.0, 9.7}; % N to E
            \node [style=new style 2] at (0, 1) {6.7, 9.7, 1.3, 0.0, 4.7}; % E to N
            \node [style=new style 2, rotate=-47] at (4.25, 5) {2.7, 7.7, 9.0, 12.7, 1.3}; % N to C
            \node [style=new style 2, rotate=-47] at (2.75, 3.5) {13.7, 3.3, 19.7, 24.3, 12.3}; % C to N
        \end{pgfonlayer}
        \begin{pgfonlayer}{edgelayer}
            		\draw [style=new edge style 0] (8.center) to (4.center); % Entail and Contradiction Top arrow
                \draw [style=new edge style 0] (14.center) to (6.center); % Entail and Neutral Top arrow
                \draw [style=new edge style 3, in=-135, out=-45, loop] (2) to ();
                \draw [style=new edge style 0] (5.center) to (13.center); %Entail and Neutral bottom arrow 
                \draw [style=new edge style 0] (3.center) to (7.center); %Entail and contra bottom arrow
                \draw [style=new edge style 0] (9.center) to (11.center); 
                \draw [style=new edge style 0] (12.center) to (10.center);
                \draw [style=new edge style 3, in=-135, out=-45, loop] (1) to ();
                \draw [style=new edge style 3, in=135, out=45, loop] (0) to ();
        \end{pgfonlayer}
    \end{tikzpicture}
    

%% file: consistency_SEMP_num_aware_GPT.tikz
    \begin{tikzpicture}[thick, scale=0.65, {every node/.style}={scale=0.65}]
        \begin{pgfonlayer}{nodelayer}
            % Nodes
            \node [style=new style 0] (0) at (0, 8) {C};
            \node [style=new style 0] (1) at (-7, 0) {E};
            \node [style=new style 0] (2) at (7, 0) {N};
            \node [style=none] (3) at (-6.5, 1.5) {};
            \node [style=none] (4) at (-6, 1) {};
            \node [style=none] (5) at (-5.75, -0.5) {};
            \node [style=none] (6) at (-5.75, 0.25) {};
            \node [style=none] (7) at (-1.25, 7.25) {};
            \node [style=none] (8) at (-0.75, 6.75) {};
            \node [style=none] (9) at (0.75, 6.75) {};
            \node [style=none] (10) at (1.25, 7.25) {};
            \node [style=none] (11) at (6, 1) {};
            \node [style=none] (12) at (6.5, 1.5) {};
            \node [style=none] (13) at (5.5, -0.5) {};
            \node [style=none] (14) at (5.5, 0.25) {};
            \node [style=new style 1] (20) at (0, 11.75) {};

            % Values
            \node [style=new style 1] at (7.5, -3.75) {21.3, 25.7, 9.3, 31.0, 19.7}; % N to N
            \node [style=new style 1] at (-7.25, -3.75) {20.0, 12.0, 4.0, 0.0, 24.0}; % E to E
            \node [style=new style 1] at (0, 11.75) {21.0, 28.7, 47.7, 29.7, 20.7}; % C to C
            \node [style=new style 2, rotate=47] at (-4.75, 4.5) {10.3, 13.3, 0.0, 0.0, 4.0}; % E to C
            \node [style=new style 2, rotate=47] at (-3, 3.25) {2.7, 2.0, 10.3, 2.7, 3.0}; % C to E
            \node [style=new style 2] at (0, -1.25) {4.7, 1.0, 1.7, 0.7, 10.7}; % N to E
            \node [style=new style 2] at (0, 1) {6.0, 8.0, 1.0, 0.0, 4.0}; % E to N
            \node [style=new style 2, rotate=-47] at (4.25, 5) {3.3, 6.0, 11.3, 13.3, 3.7}; % N to C
            \node [style=new style 2, rotate=-47] at (2.75, 3.5) {10.7, 3.3, 14.7, 22.7, 10.3}; % C to N
        \end{pgfonlayer}
        \begin{pgfonlayer}{edgelayer}
            		\draw [style=new edge style 0] (8.center) to (4.center); % Entail and Contradiction Top arrow
                \draw [style=new edge style 0] (14.center) to (6.center); % Entail and Neutral Top arrow
                \draw [style=new edge style 3, in=-135, out=-45, loop] (2) to ();
                \draw [style=new edge style 0] (5.center) to (13.center); %Entail and Neutral bottom arrow 
                \draw [style=new edge style 0] (3.center) to (7.center); %Entail and contra bottom arrow
                \draw [style=new edge style 0] (9.center) to (11.center); 
                \draw [style=new edge style 0] (12.center) to (10.center);
                \draw [style=new edge style 3, in=-135, out=-45, loop] (1) to ();
                \draw [style=new edge style 3, in=135, out=45, loop] (0) to ();
        \end{pgfonlayer}
    \end{tikzpicture}
    

%% file: consistency_SEMP_loc_aware_GPT.tikz
    \begin{tikzpicture}[thick, scale=0.65, {every node/.style}={scale=0.65}]
        \begin{pgfonlayer}{nodelayer}
            % Nodes
            \node [style=new style 0] (0) at (0, 8) {C};
            \node [style=new style 0] (1) at (-7, 0) {E};
            \node [style=new style 0] (2) at (7, 0) {N};
            \node [style=none] (3) at (-6.5, 1.5) {};
            \node [style=none] (4) at (-6, 1) {};
            \node [style=none] (5) at (-5.75, -0.5) {};
            \node [style=none] (6) at (-5.75, 0.25) {};
            \node [style=none] (7) at (-1.25, 7.25) {};
            \node [style=none] (8) at (-0.75, 6.75) {};
            \node [style=none] (9) at (0.75, 6.75) {};
            \node [style=none] (10) at (1.25, 7.25) {};
            \node [style=none] (11) at (6, 1) {};
            \node [style=none] (12) at (6.5, 1.5) {};
            \node [style=none] (13) at (5.5, -0.5) {};
            \node [style=none] (14) at (5.5, 0.25) {};
            \node [style=new style 1] (20) at (0, 11.75) {};

            % Values
            \node [style=new style 1] at (7.5, -3.75) {20.7, 24.0, 14.0, 32.0, 23.3}; % N to N
            \node [style=new style 1] at (-7.25, -3.75) {22.0, 12.7, 3.3, 0.0, 23.7}; % E to E
            \node [style=new style 1] at (0, 11.75) {17.7, 27.0, 43.7, 27.3, 17.0}; % C to C
            \node [style=new style 2, rotate=47] at (-4.75, 4.5) {7.7, 11.0, 0.0, 0.0, 3.0}; % E to C
            \node [style=new style 2, rotate=47] at (-3, 3.25) {3.0, 3.0, 9.3, 1.3, 4.3}; % C to E
            \node [style=new style 2] at (0, -1.25) {6.0, 1.3, 1.0, 0.7, 9.3}; % N to E
            \node [style=new style 2] at (0, 1) {6.7, 9.7, 1.7, 0.0, 5.3}; % E to N
            \node [style=new style 2, rotate=-47] at (4.25, 5) {2.7, 7.3, 7.3, 12.3, 1.3}; % N to C
            \node [style=new style 2, rotate=-47] at (2.75, 3.5) {13.7, 4.0, 19.7, 26.3, 12.7}; % C to N
        \end{pgfonlayer}
        \begin{pgfonlayer}{edgelayer}
            		\draw [style=new edge style 0] (8.center) to (4.center); % Entail and Contradiction Top arrow
                \draw [style=new edge style 0] (14.center) to (6.center); % Entail and Neutral Top arrow
                \draw [style=new edge style 3, in=-135, out=-45, loop] (2) to ();
                \draw [style=new edge style 0] (5.center) to (13.center); %Entail and Neutral bottom arrow 
                \draw [style=new edge style 0] (3.center) to (7.center); %Entail and contra bottom arrow
                \draw [style=new edge style 0] (9.center) to (11.center); 
                \draw [style=new edge style 0] (12.center) to (10.center);
                \draw [style=new edge style 3, in=-135, out=-45, loop] (1) to ();
                \draw [style=new edge style 3, in=135, out=45, loop] (0) to ();
        \end{pgfonlayer}
    \end{tikzpicture}
    

%% file: consistency_SEMP_stan_aware_GPT.tikz
    \begin{tikzpicture}[thick, scale=0.65, {every node/.style}={scale=0.65}]
        \begin{pgfonlayer}{nodelayer}
            % Nodes
            \node [style=new style 0] (0) at (0, 8) {C};
            \node [style=new style 0] (1) at (-7, 0) {E};
            \node [style=new style 0] (2) at (7, 0) {N};
            \node [style=none] (3) at (-6.5, 1.5) {};
            \node [style=none] (4) at (-6, 1) {};
            \node [style=none] (5) at (-5.75, -0.5) {};
            \node [style=none] (6) at (-5.75, 0.25) {};
            \node [style=none] (7) at (-1.25, 7.25) {};
            \node [style=none] (8) at (-0.75, 6.75) {};
            \node [style=none] (9) at (0.75, 6.75) {};
            \node [style=none] (10) at (1.25, 7.25) {};
            \node [style=none] (11) at (6, 1) {};
            \node [style=none] (12) at (6.5, 1.5) {};
            \node [style=none] (13) at (5.5, -0.5) {};
            \node [style=none] (14) at (5.5, 0.25) {};
            \node [style=new style 1] (20) at (0, 11.75) {};

            % Values
            \node [style=new style 1] at (7.5, -3.75) {23.0, 26.7, 13.7, 33.7, 27.3}; % N to N
            \node [style=new style 1] at (-7.25, -3.75) {23.7, 12.3, 3.3, 0.0, 23.7}; % E to E
            \node [style=new style 1] at (0, 11.75) {16.0, 26.0, 43.0, 27.7, 16.3}; % C to C
            \node [style=new style 2, rotate=47] at (-4.75, 4.5) {6.3, 10.3, 0.0, 0.0, 2.0}; % E to C
            \node [style=new style 2, rotate=47] at (-3, 3.25) {4.0, 3.0, 6.0, 1.7, 3.0}; % C to E
            \node [style=new style 2] at (0, -1.25) {5.0, 0.7, 0.3, 0.3, 6.0}; % N to E
            \node [style=new style 2] at (0, 1) {6.3, 10.7, 1.7, 0.0, 6.3}; % E to N
            \node [style=new style 2, rotate=-47] at (4.25, 5) {1.3, 5.3, 8.3, 11.0, 0.7}; % N to C
            \node [style=new style 2, rotate=-47] at (2.75, 3.5) {14.3, 5.0, 23.7, 25.7, 14.7}; % C to N
        \end{pgfonlayer}
        \begin{pgfonlayer}{edgelayer}
            		\draw [style=new edge style 0] (8.center) to (4.center); % Entail and Contradiction Top arrow
                \draw [style=new edge style 0] (14.center) to (6.center); % Entail and Neutral Top arrow
                \draw [style=new edge style 3, in=-135, out=-45, loop] (2) to ();
                \draw [style=new edge style 0] (5.center) to (13.center); %Entail and Neutral bottom arrow 
                \draw [style=new edge style 0] (3.center) to (7.center); %Entail and contra bottom arrow
                \draw [style=new edge style 0] (9.center) to (11.center); 
                \draw [style=new edge style 0] (12.center) to (10.center);
                \draw [style=new edge style 3, in=-135, out=-45, loop] (1) to ();
                \draw [style=new edge style 3, in=135, out=45, loop] (0) to ();
        \end{pgfonlayer}
    \end{tikzpicture}
    

%% file: consistency_OP_ZS_GPT.tikz
    \begin{tikzpicture}[thick, scale=0.65, {every node/.style}={scale=0.65}]
        \begin{pgfonlayer}{nodelayer}
            % Nodes
            \node [style=new style 0] (0) at (0, 8) {C};
            \node [style=new style 0] (1) at (-7, 0) {E};
            \node [style=new style 0] (2) at (7, 0) {N};
            \node [style=none] (3) at (-6.5, 1.5) {};
            \node [style=none] (4) at (-6, 1) {};
            \node [style=none] (5) at (-5.75, -0.5) {};
            \node [style=none] (6) at (-5.75, 0.25) {};
            \node [style=none] (7) at (-1.25, 7.25) {};
            \node [style=none] (8) at (-0.75, 6.75) {};
            \node [style=none] (9) at (0.75, 6.75) {};
            \node [style=none] (10) at (1.25, 7.25) {};
            \node [style=none] (11) at (6, 1) {};
            \node [style=none] (12) at (6.5, 1.5) {};
            \node [style=none] (13) at (5.5, -0.5) {};
            \node [style=none] (14) at (5.5, 0.25) {};
            \node [style=new style 1] (20) at (0, 11.75) {};

            % Values
            \node [style=new style 1] at (7.5, -3.75) {20.0, 23.7, 9.3, 21.7, 18.7}; % N to N
            \node [style=new style 1] at (-7.25, -3.75) {8.3, 4.7, 3.0, 0.0, 23.0}; % E to E
            \node [style=new style 1] at (0, 11.75) {23.0, 24.7, 50.3, 39.3, 18.7}; % C to C
            \node [style=new style 2, rotate=47] at (-4.75, 4.5) {14.0, 18.3, 0.0, 0.0, 5.3}; % E to C
            \node [style=new style 2, rotate=47] at (-3, 3.25) {0.3, 0.7, 8.3, 0.3, 4.3}; % C to E
            \node [style=new style 2] at (0, -1.25) {1.7, 0.7, 1.0, 0.0, 9.7}; % N to E
            \node [style=new style 2] at (0, 1) {14.0, 10.3, 2.0, 0.0, 3.7}; % E to N
            \node [style=new style 2, rotate=-47] at (4.25, 5) {7.7, 8.3, 12.0, 23.3, 5.7}; % N to C
            \node [style=new style 2, rotate=-47] at (2.75, 3.5) {11.0, 8.7, 14.0, 15.3, 11.0}; % C to N
        \end{pgfonlayer}
        \begin{pgfonlayer}{edgelayer}
            		\draw [style=new edge style 0] (8.center) to (4.center); % Entail and Contradiction Top arrow
                \draw [style=new edge style 0] (14.center) to (6.center); % Entail and Neutral Top arrow
                \draw [style=new edge style 3, in=-135, out=-45, loop] (2) to ();
                \draw [style=new edge style 0] (5.center) to (13.center); %Entail and Neutral bottom arrow 
                \draw [style=new edge style 0] (3.center) to (7.center); %Entail and contra bottom arrow
                \draw [style=new edge style 0] (9.center) to (11.center); 
                \draw [style=new edge style 0] (12.center) to (10.center);
                \draw [style=new edge style 3, in=-135, out=-45, loop] (1) to ();
                \draw [style=new edge style 3, in=135, out=45, loop] (0) to ();
        \end{pgfonlayer}
    \end{tikzpicture}
    

%% file: consistency_OP_COT_GPT.tikz
    \begin{tikzpicture}[thick, scale=0.65, {every node/.style}={scale=0.65}]
        \begin{pgfonlayer}{nodelayer}
            % Nodes
            \node [style=new style 0] (0) at (0, 8) {C};
            \node [style=new style 0] (1) at (-7, 0) {E};
            \node [style=new style 0] (2) at (7, 0) {N};
            \node [style=none] (3) at (-6.5, 1.5) {};
            \node [style=none] (4) at (-6, 1) {};
            \node [style=none] (5) at (-5.75, -0.5) {};
            \node [style=none] (6) at (-5.75, 0.25) {};
            \node [style=none] (7) at (-1.25, 7.25) {};
            \node [style=none] (8) at (-0.75, 6.75) {};
            \node [style=none] (9) at (0.75, 6.75) {};
            \node [style=none] (10) at (1.25, 7.25) {};
            \node [style=none] (11) at (6, 1) {};
            \node [style=none] (12) at (6.5, 1.5) {};
            \node [style=none] (13) at (5.5, -0.5) {};
            \node [style=none] (14) at (5.5, 0.25) {};
            \node [style=new style 1] (20) at (0, 11.75) {};

            % Values
            \node [style=new style 1] at (7.5, -3.75) {17.3, 24.3, 10.3, 28.0, 24.3}; % N to N
            \node [style=new style 1] at (-7.25, -3.75) {22.7, 15.7, 2.7, 0.0, 24.3}; % E to E
            \node [style=new style 1] at (0, 11.75) {23.0, 29.7, 46.3, 33.0, 19.3}; % C to C
            \node [style=new style 2, rotate=47] at (-4.75, 4.5) {9.0, 8.0, 0.0, 0.0, 3.7}; % E to C
            \node [style=new style 2, rotate=47] at (-3, 3.25) {3.3, 1.0, 8.0, 2.3, 3.7}; % C to E
            \node [style=new style 2] at (0, -1.25) {5.7, 0.7, 2.3, 0.7, 7.3}; % N to E
            \node [style=new style 2] at (0, 1) {4.7, 9.7, 2.3, 0.0, 4.0}; % E to N
            \node [style=new style 2, rotate=-47] at (4.25, 5) {6.3, 7.7, 9.7, 16.3, 2.3}; % N to C
            \node [style=new style 2, rotate=-47] at (2.75, 3.5) {8.0, 3.3, 18.3, 19.7, 11.0}; % C to N
        \end{pgfonlayer}
        \begin{pgfonlayer}{edgelayer}
            		\draw [style=new edge style 0] (8.center) to (4.center); % Entail and Contradiction Top arrow
                \draw [style=new edge style 0] (14.center) to (6.center); % Entail and Neutral Top arrow
                \draw [style=new edge style 3, in=-135, out=-45, loop] (2) to ();
                \draw [style=new edge style 0] (5.center) to (13.center); %Entail and Neutral bottom arrow 
                \draw [style=new edge style 0] (3.center) to (7.center); %Entail and contra bottom arrow
                \draw [style=new edge style 0] (9.center) to (11.center); 
                \draw [style=new edge style 0] (12.center) to (10.center);
                \draw [style=new edge style 3, in=-135, out=-45, loop] (1) to ();
                \draw [style=new edge style 3, in=135, out=45, loop] (0) to ();
        \end{pgfonlayer}
    \end{tikzpicture}
    

%% file: consistency_SEQ_column_DSC.tikz
\begin{tikzpicture}[thick, scale=0.65, {every node/.style}={scale=0.65}]
        \begin{pgfonlayer}{nodelayer}
            % Nodes
            \node [style=new style 0] (0) at (0, 8) {C};
            \node [style=new style 0] (1) at (-7, 0) {E};
            \node [style=new style 0] (2) at (7, 0) {N};
            \node [style=none] (3) at (-6.5, 1.5) {};
            \node [style=none] (4) at (-6, 1) {};
            \node [style=none] (5) at (-5.75, -0.5) {};
            \node [style=none] (6) at (-5.75, 0.25) {};
            \node [style=none] (7) at (-1.25, 7.25) {};
            \node [style=none] (8) at (-0.75, 6.75) {};
            \node [style=none] (9) at (0.75, 6.75) {};
            \node [style=none] (10) at (1.25, 7.25) {};
            \node [style=none] (11) at (6, 1) {};
            \node [style=none] (12) at (6.5, 1.5) {};
            \node [style=none] (13) at (5.5, -0.5) {};
            \node [style=none] (14) at (5.5, 0.25) {};
            \node [style=new style 1] (20) at (0, 11.75) {};

            % Values
            \node [style=new style 1] at (7.5, -3.75) {16.7, 9.3, 9.7, 30.3, 23.8}; % N to N
            \node [style=new style 1] at (-7.25, -3.75) {16.7, 10.1, 3.7, 0.0, 22.3}; % E to E
            \node [style=new style 1] at (0, 11.75) {23.1, 25.7, 46.6, 41.3, 19.7}; % C to C
            \node [style=new style 2, rotate=47] at (-4.75, 4.5) {13.8, 22.1, 0.2, 0.0, 9.1}; % E to C
            \node [style=new style 2, rotate=47] at (-3, 3.25) {7.0, 8.2, 24.6, 10.6, 7.8}; % C to E
            \node [style=new style 2] at (0, -1.25) {3.4, 4.7, 2.9, 1.0, 5.3}; % N to E
            \node [style=new style 2] at (0, 1) {4.1, 0.1, 0.2, 0.0, 2.7}; % E to N
            \node [style=new style 2, rotate=-47] at (4.25, 5) {11.6, 19.7, 11.2, 10.2, 4.9}; % N to C
            \node [style=new style 2, rotate=-47] at (2.75, 3.5) {3.6, 0.1, 0.9, 6.6, 4.4}; % C to N
        \end{pgfonlayer}
        \begin{pgfonlayer}{edgelayer}
            		\draw [style=new edge style 0] (8.center) to (4.center); % Entail and Contradiction Top arrow
                \draw [style=new edge style 0] (14.center) to (6.center); % Entail and Neutral Top arrow
                \draw [style=new edge style 3, in=-135, out=-45, loop] (2) to ();
                \draw [style=new edge style 0] (5.center) to (13.center); %Entail and Neutral bottom arrow 
                \draw [style=new edge style 0] (3.center) to (7.center); %Entail and contra bottom arrow
                \draw [style=new edge style 0] (9.center) to (11.center); 
                \draw [style=new edge style 0] (12.center) to (10.center);
                \draw [style=new edge style 3, in=-135, out=-45, loop] (1) to ();
                \draw [style=new edge style 3, in=135, out=45, loop] (0) to ();
        \end{pgfonlayer}
    \end{tikzpicture}

%% file: consistency_MIX_500.tikz
    \begin{tikzpicture}[thick, scale=0.65, {every node/.style}={scale=0.65}]
        \begin{pgfonlayer}{nodelayer}
            % Nodes
            \node [style=new style 0] (0) at (0, 8) {C};
            \node [style=new style 0] (1) at (-7, 0) {E};
            \node [style=new style 0] (2) at (7, 0) {N};
            \node [style=none] (3) at (-6.5, 1.5) {};
            \node [style=none] (4) at (-6, 1) {};
            \node [style=none] (5) at (-5.75, -0.5) {};
            \node [style=none] (6) at (-5.75, 0.25) {};
            \node [style=none] (7) at (-1.25, 7.25) {};
            \node [style=none] (8) at (-0.75, 6.75) {};
            \node [style=none] (9) at (0.75, 6.75) {};
            \node [style=none] (10) at (1.25, 7.25) {};
            \node [style=none] (11) at (6, 1) {};
            \node [style=none] (12) at (6.5, 1.5) {};
            \node [style=none] (13) at (5.5, -0.5) {};
            \node [style=none] (14) at (5.5, 0.25) {};
            \node [style=new style 1] (20) at (0, 11.75) {};

            % Values
            \node [style=new style 1] at (7.5, -3.75) {24.5, 26.3, 17.7, 33.5, 25.9}; % N to N
            \node [style=new style 1] at (-7.25, -3.75) {17.3, 17.7, 3.3, 0.0, 18.6}; % E to E
            \node [style=new style 1] at (0, 11.75) {15.2, 22.0, 61.0, 47.1, 21.1}; % C to C
            \node [style=new style 2, rotate=47] at (-4.75, 4.5) {9.1, 10.2, 0.6, 0.0, 11.5}; % E to C
            \node [style=new style 2, rotate=47] at (-3, 3.25) {7.8, 10.1, 7.4, 3.3, 4.4}; % C to E
            \node [style=new style 2] at (0, -1.25) {3.4, 3.6, 0.3, 0.2, 4.2}; % N to E
            \node [style=new style 2] at (0, 1) {8.2, 4.4, 0.1, 0.0, 4.0}; % E to N
            \node [style=new style 2, rotate=-47] at (4.25, 5) {3.9, 3.7, 5.9, 7.9, 3.9}; % N to C
            \node [style=new style 2, rotate=-47] at (2.75, 3.5) {10.7, 1.9, 3.7, 8.2, 6.4}; % C to N
        \end{pgfonlayer}
        \begin{pgfonlayer}{edgelayer}
            		\draw [style=new edge style 0] (8.center) to (4.center); % Entail and Contradiction Top arrow
                \draw [style=new edge style 0] (14.center) to (6.center); % Entail and Neutral Top arrow
                \draw [style=new edge style 3, in=-135, out=-45, loop] (2) to ();
                \draw [style=new edge style 0] (5.center) to (13.center); %Entail and Neutral bottom arrow 
                \draw [style=new edge style 0] (3.center) to (7.center); %Entail and contra bottom arrow
                \draw [style=new edge style 0] (9.center) to (11.center); 
                \draw [style=new edge style 0] (12.center) to (10.center);
                \draw [style=new edge style 3, in=-135, out=-45, loop] (1) to ();
                \draw [style=new edge style 3, in=135, out=45, loop] (0) to ();
        \end{pgfonlayer}
    \end{tikzpicture}

%% file: consistency_DYNMIX_1500.tikz
\begin{tikzpicture}[thick, scale=0.65, {every node/.style}={scale=0.65}]
        \begin{pgfonlayer}{nodelayer}
            % Nodes
            \node [style=new style 0] (0) at (0, 8) {C};
            \node [style=new style 0] (1) at (-7, 0) {E};
            \node [style=new style 0] (2) at (7, 0) {N};
            \node [style=none] (3) at (-6.5, 1.5) {};
            \node [style=none] (4) at (-6, 1) {};
            \node [style=none] (5) at (-5.75, -0.5) {};
            \node [style=none] (6) at (-5.75, 0.25) {};
            \node [style=none] (7) at (-1.25, 7.25) {};
            \node [style=none] (8) at (-0.75, 6.75) {};
            \node [style=none] (9) at (0.75, 6.75) {};
            \node [style=none] (10) at (1.25, 7.25) {};
            \node [style=none] (11) at (6, 1) {};
            \node [style=none] (12) at (6.5, 1.5) {};
            \node [style=none] (13) at (5.5, -0.5) {};
            \node [style=none] (14) at (5.5, 0.25) {};
            \node [style=new style 1] (20) at (0, 11.75) {};

            % Values
            \node [style=new style 1] at (7.5, -3.75) {22.5, 26.5, 16.6, 31.9, 25.2}; % N to N
            \node [style=new style 1] at (-7.25, -3.75) {17.7, 15.6, 3.5, 0.0, 19.6}; % E to E
            \node [style=new style 1] at (0, 11.75) {16.4, 23.4, 60.7, 47.7, 20.8}; % C to C
            \node [style=new style 2, rotate=47] at (-4.75, 4.5) {9.9, 12.0, 0.5, 0.0, 10.9}; % E to C
            \node [style=new style 2, rotate=47] at (-3, 3.25) {8.7, 9.0, 8.5, 4.0, 4.4}; % C to E
            \node [style=new style 2] at (0, -1.25) {4.0, 2.8, 0.3, 0.2, 4.3}; % N to E
            \node [style=new style 2] at (0, 1) {7.0, 4.7, 0.1, 0.0, 3.5}; % E to N
            \node [style=new style 2, rotate=-47] at (4.25, 5) {5.2, 4.4, 6.9, 9.4, 4.5}; % N to C
            \node [style=new style 2, rotate=-47] at (2.75, 3.5) {8.6, 1.7, 2.9, 6.8, 6.7}; % C to N
        \end{pgfonlayer}
        \begin{pgfonlayer}{edgelayer}
            		\draw [style=new edge style 0] (8.center) to (4.center); % Entail and Contradiction Top arrow
                \draw [style=new edge style 0] (14.center) to (6.center); % Entail and Neutral Top arrow
                \draw [style=new edge style 3, in=-135, out=-45, loop] (2) to ();
                \draw [style=new edge style 0] (5.center) to (13.center); %Entail and Neutral bottom arrow 
                \draw [style=new edge style 0] (3.center) to (7.center); %Entail and contra bottom arrow
                \draw [style=new edge style 0] (9.center) to (11.center); 
                \draw [style=new edge style 0] (12.center) to (10.center);
                \draw [style=new edge style 3, in=-135, out=-45, loop] (1) to ();
                \draw [style=new edge style 3, in=135, out=45, loop] (0) to ();
        \end{pgfonlayer}
    \end{tikzpicture}